\documentclass[11pt, a4paper, logo, twocolumn, copyright]{googledeepmind}

\usepackage[authoryear, sort&compress, round]{natbib}
\usepackage{multicol}
\usepackage{multirow}
\usepackage{subcaption}
\usepackage{tablefootnote}
\usepackage{listings}
\usepackage{tcolorbox}

\renewcommand{\cite}[1]{\citep{#1}}

\newcommand{\ib}{{\it IMO-Bench}}
\newcommand{\iab}{{\it IMO-AnswerBench}}
\newcommand{\ipb}{{\it IMO-ProofBench}}
\newcommand{\igb}{{\it IMO-GradingBench}}

\newcommand{\iabs}{IMO-AnswerBench} 
\newcommand{\ipbs}{IMO-ProofBench} 
\newcommand{\igbs}{IMO-GradingBench} 
\newcommand{\ag}{{\it AnswerAutoGrader}}
\newcommand{\pg}{{\it ProofAutoGrader}}
\newcommand{\ags}{{AnswerAutoGrader}}
\newcommand{\pgs}{{ProofAutoGrader}}

\newcommand{\opus}{Claude Opus 4} 
\newcommand{\deepseekv}{DeepSeek V3}
\newcommand{\deepseekr}{DeepSeek R1}
\newcommand{\kimi}{Kimi-K2-Instruct} 
\newcommand{\sonnet}{Claude Sonnet 4}
\newcommand{\qwen}{Qwen3-235B} 
\newcommand{\ofourmini}{o4-mini ({\it high reasoning})}
\newcommand{\grok}{Grok 4} 
\newcommand{\gemini}{Gemini 2.5 Pro}
\newcommand{\othree}{o3} 
\newcommand{\gpt}{GPT-5} 
\newcommand{\grokh}{Grok 4 ({\it heavy})}
\newcommand{\linyang}{Gemini 2.5 Pro with (Huang \& Yang, 2025)}
\newcommand{\deepthink}{Gemini 2.5 Deep Think}
\newcommand{\aero}{Gemini Deep Think ({\it IMO Gold})}
\newcommand{\imogold}{IMO-gold}

\usepackage{soul}

\definecolor{ZSBaseline}{HTML}{ff8d13}
\definecolor{KDBaseline}{HTML}{bd00ff}
\definecolor{DABaseline}{HTML}{2782ed}
\definecolor{OurColor}{HTML}{36aa70}
\definecolor{ExampleBg}{HTML}{ffffff}
\definecolor{ExampleTitle}{HTML}{545f7f}
\newmdenv[
    roundcorner=5pt,
    backgroundcolor=ExampleBg,
    linecolor=ExampleTitle,
    outerlinewidth=0.5pt,
    frametitlebackgroundcolor=ExampleTitle,
    frametitlefont={\bfseries\color{white}},
    nobreak=true,  
]{problemexample}

\AtBeginEnvironment{problemexample}{\setlength{\parindent}{0pt}}

\title{Towards Robust Mathematical Reasoning}

\author{
 Thang Luong\textsuperscript{$\diamond$},
 Dawsen Hwang\textsuperscript{*},
 Hoang H. Nguyen\textsuperscript{*$\dagger$},
 Golnaz Ghiasi\textsuperscript{*},
 Yuri Chervonyi\textsuperscript{*},
 Insuk Seo\textsuperscript{*$\dagger$},
 Junsu Kim\textsuperscript{*},
 Garrett Bingham,
 Jonathan Lee,
 Swaroop Mishra\textsuperscript{$\dagger$},
 Alex Zhai,
 Clara Huiyi Hu,
 Henryk Michalewski, Jimin Kim\textsuperscript{$\dagger$},
 Jeonghyun Ahn\textsuperscript{$\dagger$},
 Junhwi Bae\textsuperscript{$\dagger$},
 Xingyou Song,
 Trieu H. Trinh,
 Quoc V. Le,
 Junehyuk Jung\textsuperscript{$\diamond$}
\\
 \textsuperscript{$\diamond$}Corresponding authors, \textsuperscript{*}Core and equal contributors, 
 \textsuperscript{$\dagger$}Work previously conducted under Google DeepMind
}
\correspondingauthor{thangluong@google.com, junehuyk@google.com \newline External affiliations: Georgia Institute of Technology (Hoang Nguyen), Seoul National University (Insuk Seo, Junsu Kim, Jimin Kim), Microsoft (Swaroop Mishra), Massachusetts Institute of Technology (Jeonghyun Ahn, Junhwi Bae), Brown University (Junehyuk Jung).}

\begin{abstract}
\footnotesize
Finding the right north-star metrics is highly critical for advancing the mathematical reasoning capabilities of foundation models, especially given that existing evaluations are either too easy or only focus on getting correct short answers. To address these issues, we present \ib{}, a suite of advanced reasoning benchmarks, vetted by a panel of top specialists and that specifically targets the level of the International Mathematical Olympiad (IMO), the most prestigious venue for young mathematicians. \iab{} first tests models on 400 diverse Olympiad problems with verifiable short answers. \ipb{} is the next-level evaluation for proof-writing capabilities, which includes both basic and advanced IMO level problems as well as detailed grading guidelines to facilitate automatic grading. These benchmarks played a crucial role in our historic achievement of the gold-level performance at IMO 2025 with Gemini Deep Think~\cite{imo-gold}. Our model achieved 80.0\% on \iabs{} and 65.7\% on the advanced \ipbs{}, surpassing the best non-Gemini models by large margins of 6.9\% and 42.4\% respectively. We also showed that autograders built with Gemini reasoning correlate well with human evaluations and construct \igb{}, with 1000 human gradings on proofs, to enable further progress in automatic evaluation of long-form answers. We hope that \ib{} will help the community towards advancing robust mathematical reasoning and release it at {\footnotesize \url{https://imobench.github.io}}.
\end{abstract}

\begin{document}
\maketitle

\section{Introduction}
The field of artificial intelligence, particularly large language or foundation models, has demonstrated remarkable progress in mathematical reasoning capabilities.  Many popular benchmarks such as GSM8K \cite{2110.14168}, 
MATH \cite{2103.03874}, and the recently popular AIME
have approached saturation, limiting their usefulness in differentiating model performances. The problems in these datasets often rely on a limited set of techniques and do not always require the deep, multi-step reasoning needed to truly evaluate AI mathematical reasoning.  Indeed, relying on final answer matching, even in recent benchmarks such as FrontierMath \cite{glazer2024frontiermath} and Humanity's Last Exam \cite{phan2025humanity}, is not entirely reliable.  It could lead to AI systems that are good at guessing answers but do not exhibit robust reasoning.

\begin{figure}[hbt!]
    \centering
    \includegraphics[width=\linewidth]{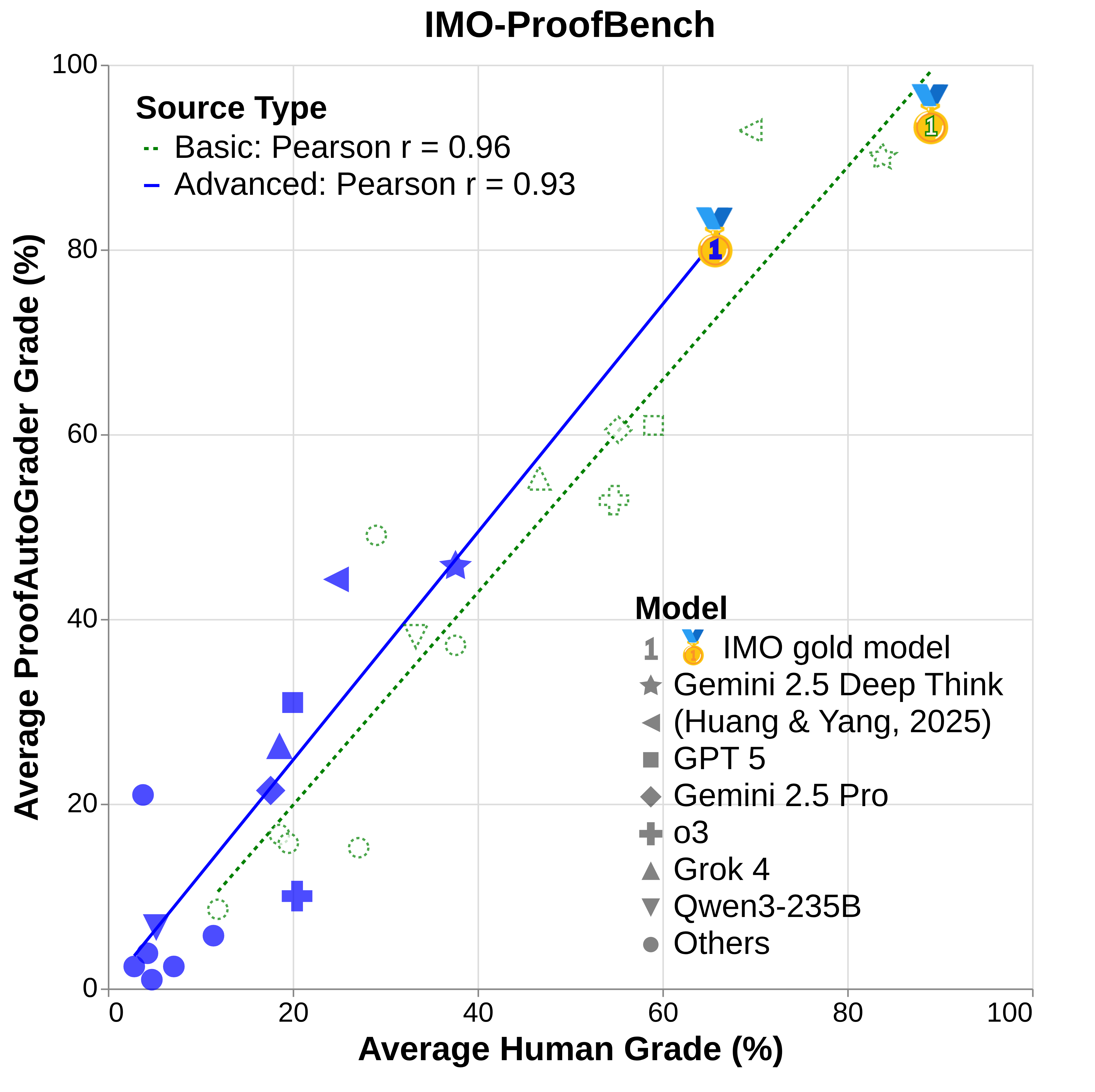}
    \caption{
    \ipb{}, a benchmark in \ib{}, for measuring proof-writing capabilities. 
        We demonstrated high correlations between human and automatic evaluations on a variety of public models, including our IMO-gold model.
        See $\S$\ref{sec:proofbench} and $\S$\ref{subsec:ipb-autograder} for more details.}
    \vspace{-5mm}
    \label{fig:autograder_external}
\end{figure}

To address these shortcomings, we propose \ib{}, a suite of benchmarks that focus on robust reasoning at the level of the International Mathematical Olympiad (IMO), the world's most celebrated arena for young mathematicians. The IMO is selected due to its notoriously difficult problems, which require not only rigorous multi-step reasoning but also a high degree of novelty, going beyond the simple application of known formulas. Such characteristics make IMO an excellent testbed for assessing reasoning capability. \ib{} covers three different tasks as summarized in Table~\ref{tab:imo_bench} and all problems were vetted by a panel of IMO medalists\footnote{Together, they won 10 gold and 5 silver IMO medals.} and mathematicians.
\begin{table}[tbh!]
\centering
\resizebox{\linewidth}{!}{%
\begin{tabular}{lrl}
\toprule
\textbf{Benchmark} & \textbf{Size} & \textbf{Task} \\
\midrule
\iab{} & 400 & Get the right answer \\
\ipb{} & 60 & Write a rigorous proof \\
\igb{} & 1000 & Grade a proof \\
\bottomrule
\end{tabular}%
}
\caption{Benchmarks in the \ib{} suite.
}
\label{tab:imo_bench}
\end{table}

\begin{figure*}[tbh!]
    \centering
    \resizebox{\linewidth}{!}{%
    \includegraphics{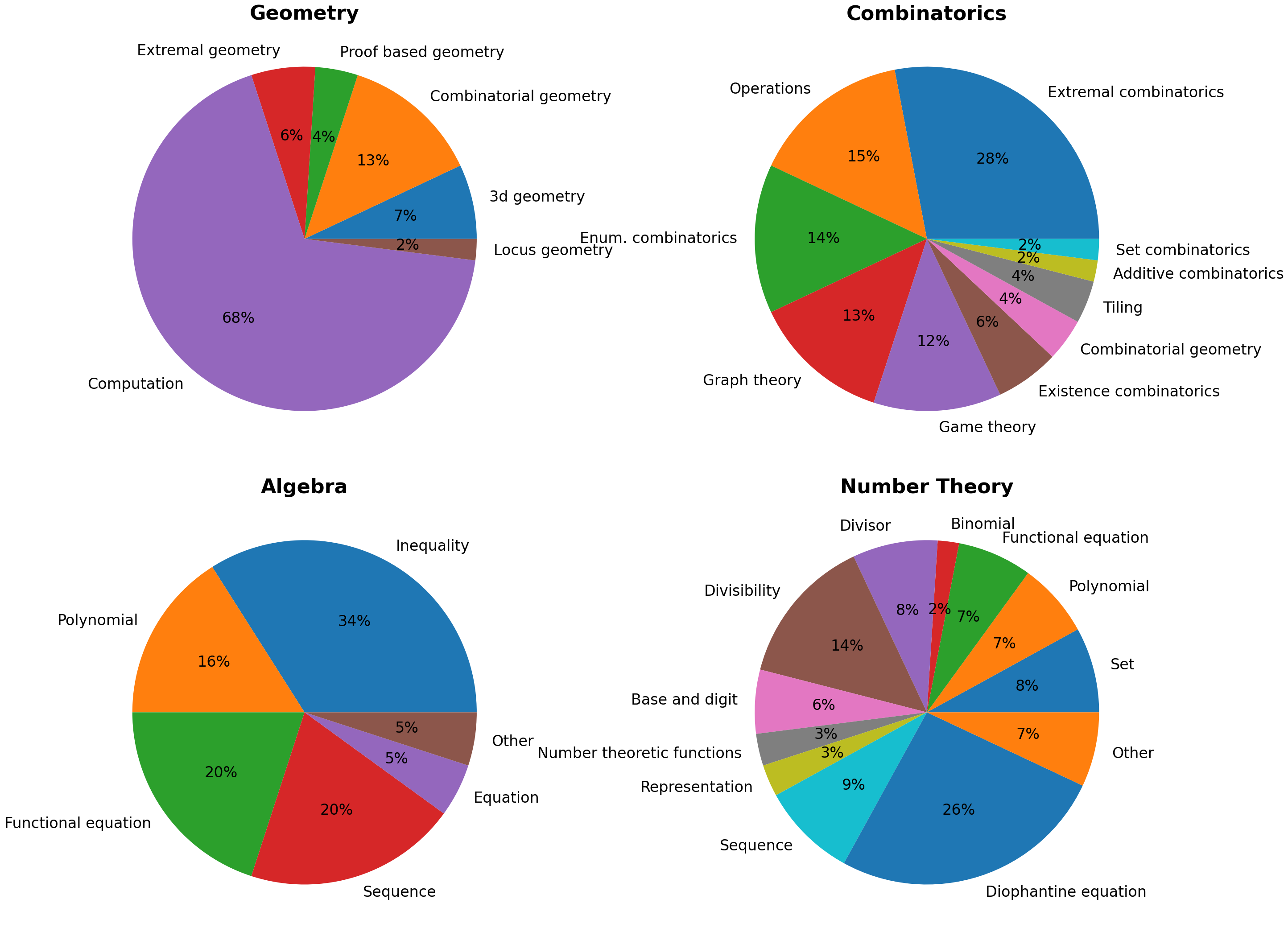}
    }
    \caption{Topic distribution by category in \iab{}. Number Theory and Combinatorics have the most topics which reflect the broad knowledge required to solve these problems while Geometry is mostly skewed towards angle and sidelength computation problems due to the nature of the short answer benchmark.}
    \label{fig:subcategories}
\end{figure*}

The first benchmark,
\iab{}, consists of 400 problems with verifiable answers carefully chosen from past Olympiad competitions and then altered by experts to avoid memorization.
Problems were chosen from a variety of topics whose solutions require different problem solving techniques to ensure a diverse representation of topics, ideas, and domain knowledge as illustrated in Figure~\ref{fig:subcategories}.

The second benchmark, \ipb{}, consists of 60 problems of varying difficulty levels, similar to those found at the IMO. While some problems have short answers, all require models to generate complete proofs. The benchmark is divided into two subsets, {\it basic} and {\it advanced}, each with 30 problems. While the basic set covers difficulty levels from pre-IMO up to IMO-Medium, problems in the advanced set are up to IMO-hard level and consist of 5 complete IMO sets, 3 of which are novel. We designed this benchmark to steer the community's focus from final answers to proofs, enabling a more rigorous assessment of AI reasoning processes. To ensure consistent evaluation, we include detailed grading schemes suitable for both human experts and automated systems. Figure~\ref{fig:autograder_external} provides an early look into the potential of automatic graders for proofs.

These two benchmarks played a crucial role in the development of our Gemini Deep Think, leading to the historic achievement of the gold-level performance at IMO 2025~\cite{imo-gold}. 
Our \imogold{} model achieved an accuracy of 80.0\% on \iabs{} by automatic evaluation, surpassing the best non-Gemini model and the best open-weight model by a large margin of 6.9\% and 19.2\% respectively. The advanced \ipbs{} is much more challenging. Our \imogold{} model scored 65.7\%, whereas the best non-Gemini and the best open-weight models performed poorly with only 23.3\% and 7.1\% accuracy according to human evaluations. Furthermore, we demonstrate that automated graders for both answers and proofs, built upon \gemini{}, achieve high correlation with expert human evaluations.

Last but not least, we introduce \igb{}, a benchmark of 1000 solutions to problems in the advanced \ipbs{}, together with grades from human experts. This resource is designed to foster progress in the automatic evaluation of long-form answers.
We release
\ib{} to the community and hope that it will spur further research towards advancing robust mathematical reasoning.

\section{\iabs{}}

\subsection{Problem Selection}
400 math problems were handpicked from various national, regional, and international  Olympiad contests, spanning across  four categories ({\it Algebra, Combinatorics, Geometry, Number Theory}). For each category, the benchmark contains 100 problems across four levels of difficulty: {\it pre-IMO} (middle school or pre-Math Olympiad problems), {\it IMO-Easy} (equivalent to Problem 1 or Problem 4 at the IMO), {\it IMO-Medium} (equivalent to Problem 2 or Problem 5 at the IMO) and {\it IMO-Hard} (equivalent to Problem 3 or Problem 6 at the IMO or post-Math Olympiad problems).  The difficulty breakdown for each category is listed in Table \ref{tab:difficulty}. 

\begin{table}[tbh!]
\centering
\resizebox{\linewidth}{!}{%
\begin{tabular}{lcccc}
\toprule
\textbf{Category} & \textbf{Pre-IMO} & \textbf{IMO-Easy} & \textbf{IMO-Medium} & \textbf{IMO-Hard} \\
\midrule
Algebra & 11 & 46 & 32 & 11 \\
Combinatorics & 4 & 19 & 31 & 46 \\
Geometry & 13 & 44 & 32 & 11 \\
Number Theory & 2 & 20 & 31 & 47 \\
\bottomrule
\end{tabular}%
}
\caption{Difficulty statistics for \iab{}.}
\label{tab:difficulty}
\end{table}

Problems with short answers were chosen so the correctness of a model's output can be quickly and reliably determined.  Given the proof-heavy nature of many math Olympiad problems, we perform an additional reformulation step for certain examples. This adjustment ensures that each problem yields a clear and nontrivial short answer, thereby reducing ambiguity during solving and verification and confirming that models utilize nontrivial reasoning. See further details in \ref{sec:consistent-eval}.

\subsection{Problem Robustification}
To avoid data memorization,
an additional step of problem modification is done via paraphrasing, changing the name of objects in the problem (such as changing point names for geometry problems), reformulating, modifying numerical values and/or adding distractors to the problem. This process is done either manually or automatically using language models. We highlight some examples in Table~\ref{tab:imo-answer-bench-examples} and detail below.

One example is an algebra problem from Austria Math Olympiad 2017. The problem is modified by making the substitution $x=a + b - c ,~ y = b + c - a$, and $z = c + a - b$ for positive real numbers $x,y,z$ with $a,~b$, and $c$ being the lengths of the sides of some triangle to obtain the modified problem in the {\it Robustified} column. This modification uses the knowledge that $a$, $b$, and $c$ are lengths of a triangle if and only if they satisfy the triangle inequalities $a+b>c$, $a+c>b$, and $b+c>a$. 

Another example is a combinatorics problem from USA TST 2005. From the original statement, the problem is modified using several techniques such as modifying numerical values (by assigning a specific value to the variable $n$ so that it is harder to guess the pattern), adding distractors (by introducing a function or variables that are not relevant to the problem), and adding a layer of challenge that could confuse the models.

Experts also reformulated original problems into equivalent ones with completely different expressions. One such example is the Czech-Slovak Math Olympiad 2017 problem. We obtained a robustified problem by transforming the governing equation and changing the objective from finding all possible values of $k$ to finding all even integers $d$ such that the number of solutions is even.

\subsection{Answer autograder} 
Even for the problems with short answers, automatic answer verification presents a few substantial challenges. The difficulty arises from two main issues: (1) ensuring that model outputs adhere to a parsable format and (2) evaluating semantically equivalent but syntactically different expressions.\footnote{For example, given the ground truth answer "$(-\infty, -4) \cup (-4, \infty)$", the answer "all real numbers except -4" should also be graded as correct.} To circumvent this issue, benchmarks such as FrontierMath \cite{glazer2024frontiermath} select problems with only numerical answers or mathematical objects that can be expressed as SymPy objects. However, this approach narrows the scope of evaluable problems and reduces robustness of the benchmark to minor formatting or syntax errors.

To address these limitations, we use large language models as automated verifiers for model answers on \iab{}. We name this approach, \ag{}, which is built by prompting the public \gemini{} model to extract final answers from generated solutions and assess their correctness against ground truths (See \ref{subsec:answergrader-prompt} for the full prompt). This method allows much more flexibility in acceptable answer formats and improves the overall robustness of our benchmark. As we demonstrate in Section~\ref{subsec:iab-automatic-eval}, \ag's performance is nearly identical to that of human evaluators, validating its use for future public usage and also for reporting the results in this work.

\section{Going Beyond Short Answers with \ipbs{}}
\label{sec:proofbench}
While the final answer accuracy provided by \iab{} offers a valuable metric for measuring mathematical abilities, it is insufficient for a comprehensive assessment of mathematical reasoning.  A final answer can be correct while the full solution contains flawed reasoning. Furthermore, many IMO-level competition problems do not come with a final short answer. Even in cases where a short answer exists, guessing the correct short answer is often significantly easier than rigorously deriving the solution.

\ipb{} is designed to evaluate the ability of AI models to construct comprehensive and valid mathematical arguments. This benchmark consists of 60 proof-based problems, curated to mirror the kinds of problems found in the IMO. While some problems may have concise numerical answers, models are only given credit if they produce correct and relevant reasoning steps. This benchmark is essential for assessing an AI's underlying reasoning process, its ability to apply mathematical principles, and its capacity to formulate coherent and logical arguments.  

\subsection{Benchmark setup}
The benchmark is divided into two subsets: a \textit{basic} set covering pre-IMO to IMO-Medium difficulty levels, and an \textit{advanced} set featuring novel, highly challenging problems simulating complete IMO examinations, up to IMO-Hard level.

The basic problem set primarily consists of rephrased versions of existing problems. Since standard IMO problems may be too challenging for most of current models, the basic set is designed to assess models in their early stages of development.  Sufficiently strong performance on the basic set would justify progression to the advanced set.
The advanced problem set features 30 problems in the style and difficulty of the IMO. The collection includes 18 novel problems crafted by IMO medalists, alongside 12 problems from recent top-tier competitions: 6 robustified from IMO 2024 and 6 directly from USAMO 2025. Table~\ref{tab:imo2024-robustified-example} provides examples of such robustified problems.

\ipb{} uses an evaluation framework designed for both simplicity and precision. We provide a primary grading guideline with four ratings ({\it Correct, Almost, Partial, Incorrect}) as detailed in Table~\ref{tab:imo_ratings}.
While this rubric offers a clear and consistent baseline, we do not restrict our expert evaluators to these four values. To allow for more nuanced assessments, human experts are empowered to use their own judgments to assign any integer score from 0 to 7 for each problem.

\begin{table}[tbh!]
\centering
\resizebox{\linewidth}{!}{%
\begin{tabular}{lcl}
\toprule
\textbf{Category} & \textbf{IMO Points} & \textbf{Solution quality} \\
\midrule
Correct & 7 & Fully correct, rigorous, and complete \\
Almost & 6 & Almost correct, minor errors \\
Partial & 1 & Mostly incorrect, some relevant results \\
Incorrect & 0 & Completely incorrect or irrelevant. \\
\bottomrule
\end{tabular}%
}
\caption{Our simplified IMO ratings.}
\label{tab:imo_ratings}
\end{table}

\subsection{Proof Autograder}\label{subsec:autorater}
 While human expert evaluation remains the gold standard for mathematical proofs, its cost and time intensity limit scalable research. To address this, we built \pg{}, an automatic grader for  \ipb{}. The autograder leverages \gemini{}, providing it with a prompt containing the problem statement, the candidate solution, a reference solution, and specific grading guidelines (see Appendix \ref{subsec:autograder-prompt}).

Automatic evaluation for informal proofs is a highly intricate task, and current systems are not yet a perfect substitute for human experts-a key distinction from \ag , whose purpose is primarily format matching. For this reason, all primary results in this paper are based on expert human evaluation to ensure all results are absolutely correct. Nevertheless, as we demonstrate in Section \ref{subsec:ipb-autograder}, we prove our autograder can be a reasonable proxy,  establishing it as a reasonable tool for the community to assess future models on \ipb{}.

\section{\igbs{}}
\label{sec:igb_methodology}

\begin{table*}[tbh!]
\centering
\resizebox{\linewidth}{!}{%
\begin{tabular}{l|c|cccc|c}
\toprule
\textbf{Model} & \textbf{Query date} & \textbf{Algebra} & \textbf{Combinatorics} & \textbf{Geometry} & \textbf{Number Theory} & \textbf{Overall} \\
\midrule
\opus{} & 2025-08-04 & 19.4\% & 20.0\% & 23.3\% & 26.6\% & 22.3\% \\
\sonnet{} & 2025-08-06 & 20.6\% & 17.8\% & 26.0\% & 27.6\% & 23.0\% \\
\deepseekv{} & 2025-09-17 & 39.0\% & 26.0\%& 35.0\%& 48.0\%& 37.0\% \\
\kimi{} & 2025-09-17 & 45.6\% & 31.1\% & 49.3\% & 56.9\% & 45.8\% \\
\qwen{} & 2025-08-20 & 57.6\% & 37.5\% & 57.6\% & 62.3\% & 53.8\% \\
\deepseekr{} & 2025-09-17 & 65.0\%& 40.0\%& 73.0\%& 65.0\%& 60.8\% \\
\othree{} & 2025-08-04 & 62.8\% & 43.0\% & 70.6\% & 68.0\% & 61.1\% \\
\gpt{} & 2025-09-17 & 69.9\%&  46.4\% & 74.8\% & 71.2\% & 65.6\% \\
\ofourmini{} & 2025-08-04 & 71.3\% & 46.6\% & 78.4\% & 75.3\% & 67.9\% \\
\gemini{} & 2025-08-04 & 73.4\% & 48.0\% & 74.3\% & 77.1\% & 68.2\% \\
\deepthink{} & 2025-08-20 & 78.0\% & {49.0\%} & {83.0\%} & {77.0\%} & {71.8\%} \\
\grok{} & 2025-08-06 & {75.5\%} & {55.9\%} & {80.1\%} & \textbf{80.9\%} & {73.1\%} \\
\aero{} & 2025-09-17 & \textbf{85.0\%} & \textbf{69.0\%} & \textbf{88.0\%} & {78.0\%} & \textbf{80.0\%}\\
\bottomrule
\end{tabular}%
}
\caption{Model accuracy on \iab{}. Results are averaged over 8 runs, except for \deepthink{} and \aero{} (single run). An evaluation of \grokh{} on 2025-08-13 using multiple paid accounts was aborted due to significant instability (only 117/400 responses were received despite multiple, hour-long attempts), and thus its results are not reported.}
\label{tab:imo-answer-bench-result}
\end{table*}

While \ipb{} evaluates proof-writing abilities, it is equally important to assess models in terms of their ability to evaluate the correctness of given solutions. This capability is crucial for developing reliable automated grading systems and improving general mathematical reasoning. 

\begin{figure}[tbh!]
    \centering
    \resizebox{\columnwidth}{!}{%
    \includegraphics{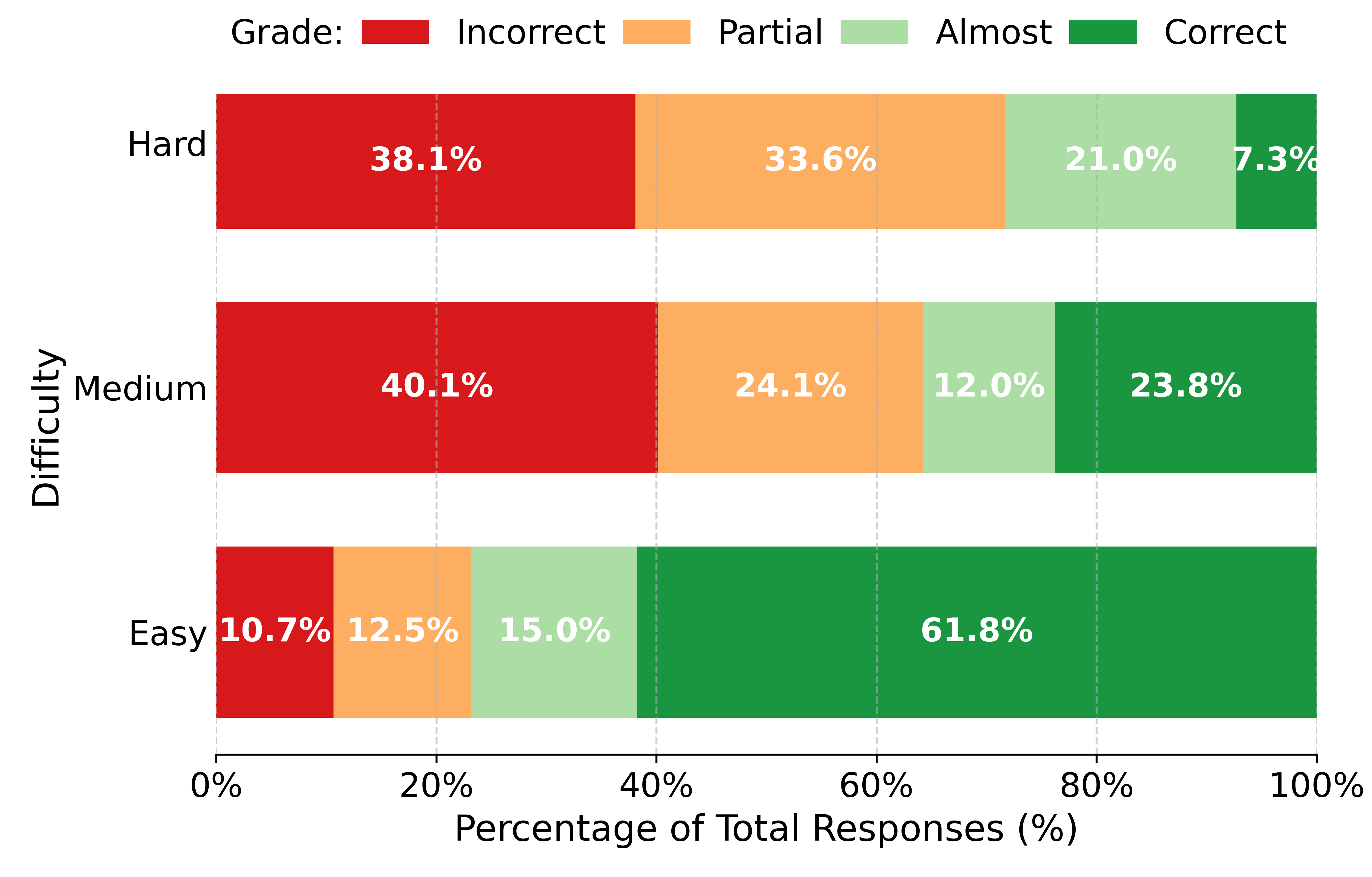}
    }
    \caption{Grade distribution for solutions in \igb{} by difficulty levels (IMO-Hard, IMO-Medium, IMO-Easy).}
    \label{fig:score-distribution-difficulty}
\end{figure}

As part of our IMO effort~\cite{imo-gold}, we have benchmarked extensively many internal models on the advanced set of \ipb{} using human evaluations.
These human gradings led to the creation of \igb{} with 1000 examples, each containing a problem statement, a proposed solution, and its human-assigned grade (on a 0–7 scale). To reduce noise from fine-grained scoring, we frame the evaluation as a four-way classification by mapping the given IMO points to the labels ({\it Correct, Almost, Partial, Incorrect}) as detailed in Table~\ref{tab:imo_ratings}. 
To ensure a robust evaluation, the dataset has been balanced with a roughly equal number of examples per category. Figure~\ref{fig:score-distribution-difficulty} illustrates that when problems are grouped by their IMO difficulties, a clear trend emerges. The proportion of correct and almost solutions decreases as the intended difficulty moves from IMO-easy to IMO-hard, while the proportion of incorrect and partial solutions increases. This confirms that the grading distribution of \igb{} aligns with its assigned difficulty levels. See further discussions in Section~\ref{sec:grade_distribution}.

\section{Results}

\begin{table}[tbh!]
\centering
\renewcommand{\arraystretch}{1.2}
\resizebox{0.9\linewidth}{!}{%
\begin{tabular}{cc|cc}
                               &   & \multicolumn{2}{c}{\textbf{\ag{}}} \\
                               &   & \textbf{0}         & \textbf{1}        \\ \hline
\multirow{2}{*}{\begin{tabular}{@{}c@{}}\textbf{Human}\\\textbf{Grade}\end{tabular}} & \textbf{0} & 274 {\footnotesize (99.6\%)} & 1 {\footnotesize (0.4\%)} \\
                               & \textbf{1} & 8 {\footnotesize (1.5\%)}   & 517 {\footnotesize (98.5\%)} \\
\end{tabular}
}
\caption{\ag{} predictions against human grades for \iab{}. 
The solutions were generated by \gemini{} and \othree{}. 
}
\label{tab:answergrader-confusion-mat}
\end{table}

\begin{table*}
\centering
\resizebox{\linewidth}{!}{%
\begin{tabular}{l|c|l|l|lcc}
\toprule
\multirow{2}{*}{\textbf{Model}} & \multirow{2}{*}{\textbf{Query date}} & \multicolumn{2}{c|}{\textbf{\ipbs{}}} & \multicolumn{3}{c}{\textbf{{\it Advanced} \ipbs{} Breakdown}} \\ 
\cmidrule(lr){3-4} \cmidrule(lr){5-7} 
& & \multicolumn{1}{c|}{\textit{Basic}} & \multicolumn{1}{c|}{\textit{Advanced}} & \multicolumn{1}{c|}{\textit{Novel}} &\multicolumn{1}{c|}{ \textit{IMO 2024$^\dagger$}} & \multicolumn{1}{c}{\textit{USAMO 2025}} \\
\midrule
\textbf{Number of Problems} &  & \multicolumn{1}{c|}{30} & \multicolumn{1}{c|}{30} & \multicolumn{1}{c|}{18} & \multicolumn{1}{c|}{6}& \multicolumn{1}{c}{6} \\
\midrule
\opus{} & 2025-08-04 &  11.9\%  & \hphantom{0}2.9\% &  \hphantom{0}0.0\% &  \hphantom{0}2.4\% & 11.9\%\\
\deepseekv{} & 2025-09-16 &  18.6\% & \hphantom{0}4.3\% &  \hphantom{0}6.3\% &  \hphantom{0}2.4\% &  \hphantom{0}0.0\% \\  
\kimi{} & 2025-08-21 & 19.5\% & \hphantom{0}7.1\% & \hphantom{0}4.0\% &  \hphantom{0}2.4\% & 21.4\% \\
\sonnet{} & 2025-09-17 & 27.1\%$^{\S1}$ & \hphantom{0}4.8\%$^{\S1}$ &  \hphantom{0}6.4\%$^{\S1}$&  \hphantom{0}2.4\% &  \hphantom{0}2.4\% \\
\deepseekr{} & 2025-09-16 &  29.0\% & \hphantom{0}3.8\% & \hphantom{0}6.4\% &  \hphantom{0}0.0\% &  \hphantom{0}0.0\% \\
\qwen{} & 2025-08-21 & 33.3\% & \hphantom{0}5.2\% &  \hphantom{0}7.1\% &  \hphantom{0}0.0\% &  \hphantom{0}4.8\% \\
\ofourmini{} & 2025-08-04 & 37.6\% & 11.4\% &  \hphantom{0}8.7\% &  \hphantom{0}7.1\% & 23.8\% \\
\grok{} & 2025-08-20 & 46.7\% & 18.6\% & 17.5\% & 16.7\% & 23.8\% \\
\othree{} & 2025-08-04 & 54.8\% & 20.5\%  & 15.1\% &  \hphantom{0}4.8\% & 52.4\% \\ 
\gemini{} & 2025-08-04 & 55.2\% & 17.6\% & 15.9\% &  \hphantom{0}7.1\% & 33.3\% \\
\gpt{} & 2025-09-18 & 59.0\% & 20.0\% & 15.9\% & 33.3\% & 19.0\%  \\
\grokh{} & 2025-07-12  & \hphantom{0}{\it NA}$^\ddagger$ & 23.3\%$^{\S3}$  & 11.1\%$^{\S3}$ & \hphantom{0}7.1\% & \textbf{76.2\%} \\
\linyang{} & 2025-07-14 & 69.5\% & 24.8\% & 17.5\% & 19.1\% & 52.4\% \\
\deepthink{} & 2025-08-20 & 83.8\% & 37.6\% & 31.7\% & 40.5\% & 52.4\% \\
\aero{} & 2025-08-02 & {\bf 89.0\%}  & {\bf 65.7\%}  & \textbf{61.1\%} & \textbf{76.2\%} & 69.0\% \\
\bottomrule
\end{tabular}%
}
\caption{
Expert evaluation results on the Basic and Advanced subsets of \ipb{}. Scores are presented as a percentage of the total possible points for the problems in each respective subset, with each problem graded from 0--7 (as described in Section \ref{sec:proof_evaluation}). The Advanced \ipb{} is further broken down by problem source. 
\\
\small{\it $^\dagger$Robustified IMO 2024 problem set, see Section~\ref{sec:proofbench}. $^\ddagger$An attempt to query \grokh{} on 2025-08-13 was unsuccessful due to model instability (only 5 of 30 problems responded with 3 attempts). $^{\S k}$Scores indicate that there were $k$ problems that were treated as incorrect (a score of 0) because of query failures (for at least 3 times).}
}
\label{tab:imo-proof-bench-manual-result}
\end{table*}

We evaluate \ib{} on a wide variety of publicly available models: \opus{} (20250514), \sonnet{}~\cite{anthropic:2025:claude4}, \deepseekv{}~\cite{deepseek:2025:v3}, \deepseekr{}~\cite{deepseek:2025:r1}, \kimi{}~\cite{moonshotai:2025:kimik2}, \qwen{} (A22B-Instruct-2507- tput)~\cite{qwen:2025:qwen3}, \othree{} (2025-04-16), \ofourmini{}~\cite{openai:2025:o3ando4mini}, \gpt{} (2025-08-07)~\cite{openai:2025:gpt5}, \gemini{}~\cite{deepmind:gemini2p5po}, \deepthink{}~\cite{deepmind:gemini2p5deepthink}, \aero{}~\cite{imo-gold}, \linyang{}~\cite{huang2025gemini25procapable}, \grok{} (0709)~\cite{xai:grok4}.
Since \linyang{} is an agentic framework rather than a single model call, Appendix \ref{appendix:linyang} contains further implementation details.

\subsection{\iabs{} with \ags{}} \label{subsec:iab-automatic-eval}

Results for \iab{} are summarized 
in Table \ref{tab:imo-answer-bench-result}.  Accuracy was determined by \ag{}, which extracts final answers from model responses and assesses their semantic equivalence to the ground truths. Our \aero{} model achieved an overall accuracy of 80.0\%, surpassing the best non-Gemini model (\grok{}) by 6.9\% and the best open-weight model (\deepseekr{}) by 19.2\%. Latest models such as \kimi{} and \gpt{} are still struggling with overall accuracy of only 45.8\% and 65.6\% respectively. 

Across the four categories of Algebra, Combinatorics, Geometry, and Number Theory, models generally perform the worst in Combinatorics, potentially highlighting difficulties with advanced abstract reasoning.
We also analyze the performances of models on the original problems, before robustification, summarized in Table~\ref{tab:imo-answer-bench-original-result}.  As anticipated, we find robustification leads to a consistent drop in performance across all models.

Lastly, we validate the reliability of \ag{} by comparing it with expert human labels. As reported in Table \ref{tab:answergrader-confusion-mat}, the autograder shows nearly perfect performance, achieving overall accuracy of 98.9\% on the positive (correct) class.

\subsection{\ipbs{} with Expert Evaluations}

Model outputs on \ipb{} were graded by human experts according to the guidelines described in Section \ref{sec:proof_evaluation}. Table \ref{tab:imo-proof-bench-manual-result} presents the results of this evaluation. Performance on the basic \ipb{} varies significantly; while most models score below 60\%, \aero{} achieves a high score of 89.0\%. The performances of other frontier models such as \qwen{} (33.3\%) and \gpt{} (59.0\%) show there is still considerable room for improvements.

The advanced \ipb{} proves to be a more significant challenge that all non-Gemini models score below 25\%.
Our \aero{} model achieved a score of 65.7\%, surpassing the best non-Gemini model (\grokh{}) by a large margin of 42.4\%. This represents a substantial leap in capability, but its distance from a perfect score indicates that even the strongest models have room for growth in sophisticated mathematical reasoning.

A breakdown of the advanced \ipb{} reveals a significant performance disparity across problem types, suggesting potential overfitting in certain models. This trend is most evident with \grokh{}, which scores 76.2\% on USAMO 2025 but only 11.1\% on novel problems. Other models, including \othree{} (52.4\% vs. 15.1\%) and \linyang{} (52.4\% vs. 17.5\%), exhibit a similar, pronounced gap.
In contrast, \aero{} scored 69.0\% on the USAMO and 61.1\% on the novel sets, indicating it has more general capabilities \cite{deepmind:gemini2p5deepthink} without overfitting to a particular dataset.
The low performances of latest frontier models such as \gpt{} and \grokh{} on the advanced \ipb{} underscore the difficulty of advanced mathematical reasoning and highlight the importance of rigorous examination the full details of model outputs for a complete understanding of their mathematical abilities.

\subsection{Autograder for \ipbs{}} \label{subsec:ipb-autograder}
To assess the feasibility of using automatic graders for proofs, we apply \pg{} to the 14 public models (Table \ref{tab:imo-proof-bench-manual-result}), which were previously graded by human experts on \ipb{}.
Figure~\ref{fig:autograder_external} shows that the average grades from \pg{} highly correlate with human grades, yielding high Pearson correlation coefficients of $0.96$ and $0.93$ on both basic and advanced problems respectively.

\begin{figure}[tbh!]
    \centering
    \resizebox{\linewidth}{!}{
    \includegraphics{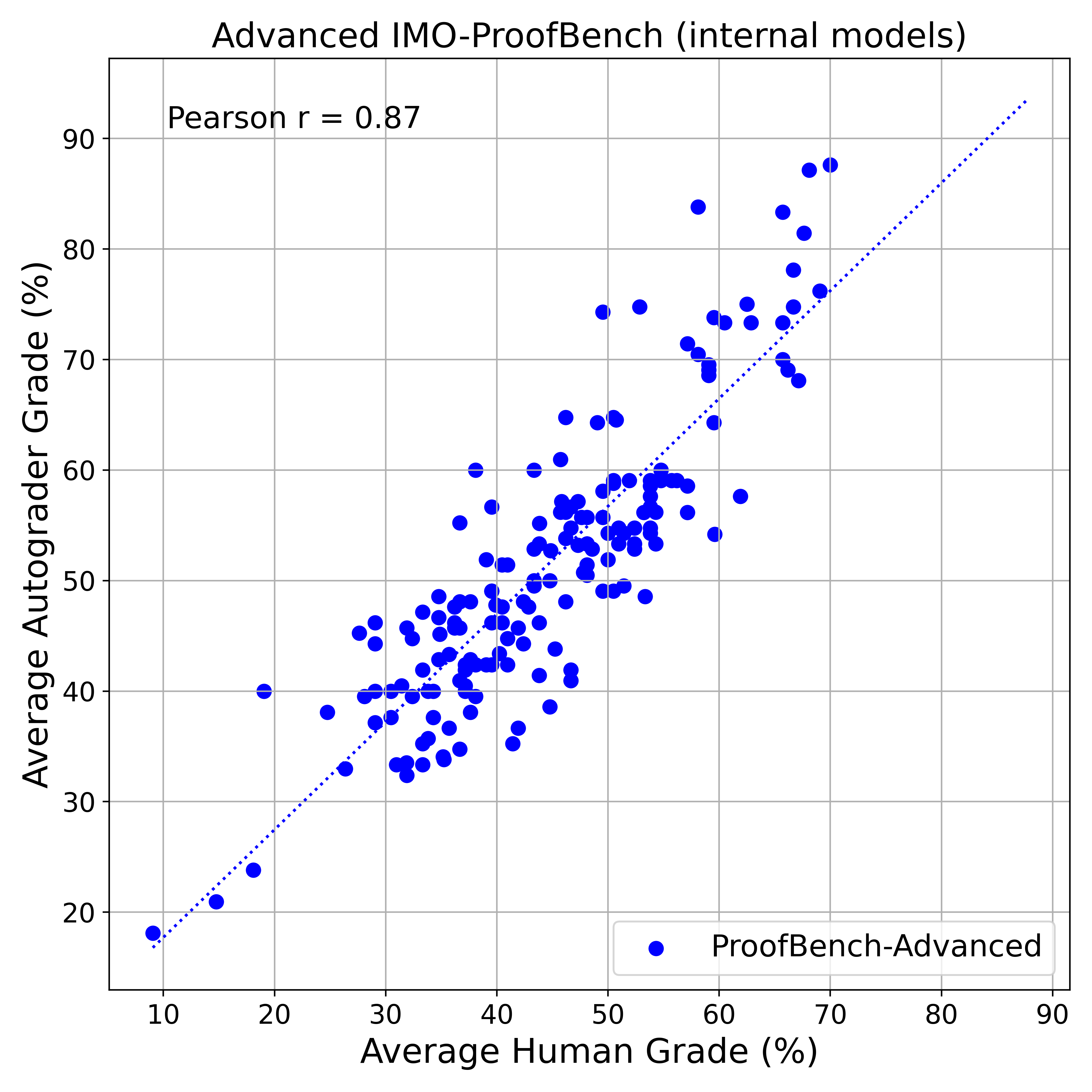}
    }
    \caption{
    Correlation between \pg{} and human experts on the advanced \ipb{}, evaluated over 170 internal models on our IMO-gold journey.
    }
    \label{fig:autograder}
\end{figure}

In addition, we also visualized, in Figure~\ref{fig:autograder}, the performance of \pg{} on 170 internal systems, developed as part of our IMO effort~\cite{imo-gold}. On this larger pool, our automatic grader  achieved a lower, but still reasonable Pearson correlation coefficient of $0.87$.

\begin{figure}[tbh!]
    \centering
    \resizebox{0.9\columnwidth}{!}{
    \includegraphics{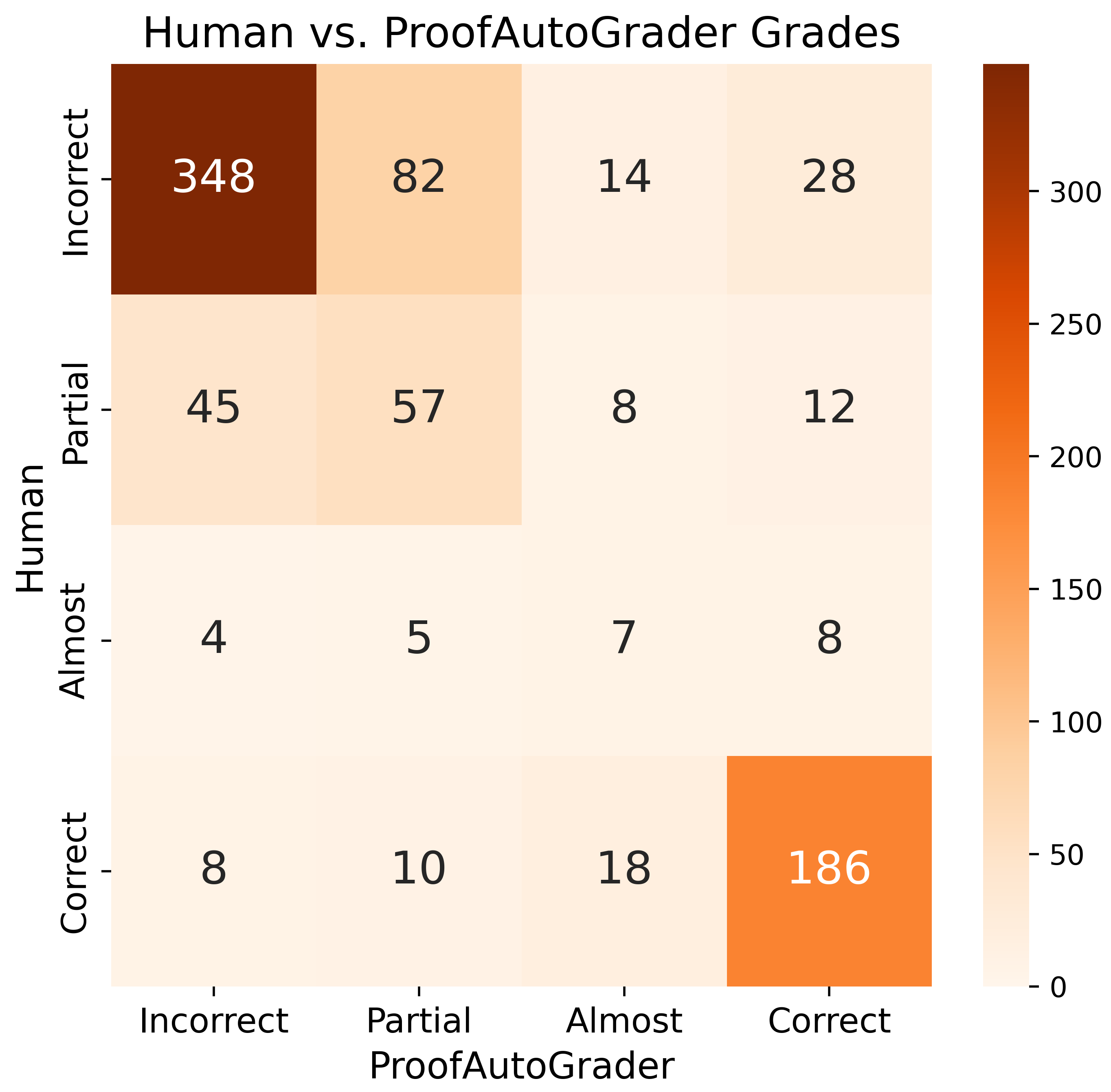}
    }
    \caption{Confusion matrix for \pg{} vs. human expert grades, over 840 solutions generated by 14 public models (See Table~\ref{tab:imo-proof-bench-manual-result}).
    }
    \label{fig:autograder-confusion-mat}
\end{figure}

To better understand the grading agreement, we visualize, in Figure \ref{fig:autograder-confusion-mat}, the confusion matrix of all human and automatic gradings on the 14 public models (for a total of 840 model solutions). We observed that most common misclassifications happened between the {\it Incorrect} and {\it Partial} classes. Overall, \pg{} shows reasonable performance, exhibiting high correlation with human experts, and also shows potential in identifying nuances that might be overlooked by human graders.

On the other hand, detailed analysis with per-solution breakdowns further reveals that
\pg{} occasionally still has weaknesses such as failures to identify high-level
logical errors or being overly punitive for unconventional yet
correct solutions. Specific examples are highlighted in appendix~\ref{sec:proofgrader-limitations}.
Therefore, while we hope that \pg{} can serve as a valuable tool for the community to evaluate models on \ipb{}, we recommend that it augments human verification to guarantee the accuracy of individual grading results.

\subsection{\igbs{}}\label{subsec:igb_results}
The \igb{} measures the ability of models in assessing the quality of a proof when provided with only problem statements and model-generated solutions, without any reference solutions or specific grading guidelines. We measure model performances under two metrics:
\begin{enumerate}
    \item {\it Accuracy} -- human gradings on a 7-point scale are first converted to 4 categories {\it (Correct, Almost, Partial, Incorrect)} corresponding to 4 buckets (7, 6-4, 3-1, 0). The categorized human gradings are then compared with model-predicted categories.
    \item {\it Mean Absolute Error (MAE)} -- model-predicted categories are converted from {\it (Correct, Almost, Partial, Incorrect)} to IMO scores (7, 6, 1, 0) according to Table~\ref{tab:imo_ratings}. We then compare with human grading ground truths on a 7-point scale.
\end{enumerate}   

\begin{table}[tbh!]
\centering
\resizebox{\linewidth}{!}{%
\begin{tabular}{l|c|c}
\toprule
\textbf{Model} & \textbf{Accuracy $\uparrow$} & \textbf{MAE $\downarrow$} \\ 
\midrule
\gemini{} & 44.3\% & 30.2\% \\
\ofourmini{} & 47.3\% & 25.2\%  \\
\deepthink{} & 52.5\% & 20.5\% \\
\othree{} & {\bf 54.0\%} & 20.2\%  \\
\aero{} & 50.2\% & {\bf 18.4\%}  \\
\bottomrule
\end{tabular}
}
\caption{\igb{} results in accuracy (higher is better) and MAE (lower is better).
}
\label{tab:imo-grading-bench-result}
\end{table}

Results for \igb{} are summarized in Table \ref{tab:imo-grading-bench-result}. In terms of accuracies, \othree{} achieved the highest performance of 54.0\%. The low accuracies highlight the fact that this benchmark is quite challenging in predicting precise categories. The MAE accounts for the fact that different categories are closer semantically, e.g., {\it Correct} vs. {\it Almost} and {\it Partial} vs. {\it Incorrect}. On this metric, \aero{} achieved the best MAE score of $18.4\%$, indicating that there is still significant room for improvement\footnote{Because of our simplified gradings (7, 6, 1, 0), the best possible grader will achieve a {\it golden} MAE of 3.9\% on \igb{}, instead of 0\%.}.

\paragraph{\textit{Comparison with \pgs{}}} Model performances on \igb{} are notably worse than what might be expected from the strong performance of \pg{}, in terms of Pearson correlation coefficients as reported in Section~\ref{subsec:ipb-autograder}. This discrepancy stems from two key methodological distinctions.
\begin{enumerate}
    \item First, \pg{} performance was measured on scores aggregated over 30 problems, which smooths out noise from individual grading variations, unlike the per-instance evaluation of \igb{}. 
    \item Second, the \igb{} evaluation provides models with minimal context—only the problem and the proposed solution; whereas for \pg{} on \ipb{}, we additionally provide both reference solutions and grading guidelines.
\end{enumerate}

These distinctions explain why \igb{}  with per-instance, minimal-context evaluation is a challenging benchmark; whereas aggregated assessments by \pg{} on \ipb{} can still yield robust model rankings.

\section{Related Work}
In recent years, harder reasoning math benchmarks have been proposed as performance on existing benchmarks becomes saturated. For example, Olympiad Bench \cite{he2024olympiadbench} and Omni-MATH \cite{gao2024omni} contain questions at the Olympiad level across diverse domains, while Humanity's Last Exam (HLE) \cite{phan2025humanity} evaluates knowledge across many domains. Other benchmarks include Brainteaser \cite{simeng-brainteaser}, which consists of long-form brainteaser puzzles, and Frontier Math~\cite{glazer2024frontiermath}, which contains hard math questions and a hidden evaluation set. MiniF2F~\cite{zheng2021minif2f} provides a benchmark for evaluating formal proofs around Olympiad-level difficulty. Reward Bench~\cite{lambert2024rewardbench} provides a benchmark to evaluate reward models. HARDMath~\cite{fan2024hardmath} presents a challenging math benchmark containing applied mathematics problems that require analytical approximation techniques. The AlphaGeometry papers \cite{Trinh2024-alphageometry-1, chervonyi2025goldmedalistperformancesolvingolympiad-alphageometry2} provide benchmarks of $80$ IMO and IMO Shortlist Euclidean geometry problems from $2000$ to $2024$, written in a domain-specific language. In contrast, \ib{} provides a suite for evaluating advanced mathematical reasoning with short answer matching and rigorous proof evaluation in natural language across a wide variety of Math Olympiad areas.

As performance on math benchmarks continues to improve, robustness benchmarks have been introduced to evaluate potential overfitting and obtain better estimates of models' true reasoning capabilities.  These benchmarks have shown that simply perturbing benchmark questions is enough to significantly hurt performance compared to the original problems.  SVAMP~\cite{patel2021nlp} generated a perturbed benchmark for word math problems, whereas Lila~\cite{mishra2022lila} contained perturbations across a diverse range of reasoning questions. The functional variant of the MATH benchmark~\cite{srivastava2024functional} demonstrated large performance drops across models when varying existing problems. Putnam-AXIOM~\cite{gulati2024putnam} similarly shows that perturbing Putnam questions causes a significant drop in model performance. MATH-Perturb~\cite{huang2025math} also adds simple perturbations to math questions~\cite{2103.03874}, and shows model performance drops, raising concerns about memorization. \citet{lightman2024verifystepbystep} propose an alternative strategy to improve model robustness by supervising the reasoning process from start to finish, rather than solely on the final outcome. This approach led to improved performance on the MATH dataset. \ib{} contributes to robust mathematical reasoning with already modified questions in \iab{}, rigorous proof requirements in \ipb{}, and the task of proof grading in \igb{}.

\section{Conclusion}
This paper introduced \ib{}, a comprehensive suite of benchmarks for robust evaluation of mathematical reasoning capabilities, including \iab{} for short answer matching, \ipb{} for full proof correctness, and \igb{} for proof verification.
The three benchmarks demonstrated that frontier models struggle on \ib{} problems and that getting the short answers right does not necessarily equate to correct mathematical reasoning for most models. 

Furthermore, we have developed and validated automated graders for both answers and proofs. Our \ag{} achieves near-human accuracy (98.9\%) , while  \pg{} shows a strong correlation (0.93-0.96 \%) with expert human scores. These tools along with \igb{} provide a scalable and reliable method for the community to evaluate future models, even as human expertise remains the gold standard for high-stakes evaluation.

By releasing \ib{}\footnote{\url{https://imobench.github.io}} to the research community, we aim to shift the community's focus from mere answer-getting to the development of deep, verifiable, and robust reasoning processes. We hope this suite will serve as a valuable tool to measure and drive progress toward more advanced and reliable artificial intelligence.

\section*{Acknowledgments}
Special thanks to Miroslav Olšák, Seongbin Jeon, Donghyun Kim, Jiwon Kang, Chu-Lan Kao, Sara Javanmardi, and Mahan Malihi for help with \ib{}.
In addition, we would like to thank Orhan Firat, Tania Bedrax-Weiss, and Ed Chi for reviewing the work and Koray Kavukcuoglu for guidance on the release of \ib{}. Last but not least, we thank all our collaborators in the IMO 2025 effort\footnote{\url{https://goo.gle/imo-gold}} for trusting \ib{} as north-star metrics along the way.

\bibliography{arxiv_latest}

\appendix

\newpage

\onecolumn
\section*{Limitations}
Our work has two primary limitations: evaluation cost and the risk of data contamination.

\textit{Evaluation Cost.} While our
automatic
grader, \pg{}, correlates strongly with human scores, it is not a perfect substitute and can introduce noise. Consequently, definitive assessments still require verification by human experts, who are both costly and difficult to source.

\textit{Future Data Contamination.}
The second limitation is the risk of long-term data contamination.
As \ib{} is publicly released, its problems and solutions will likely be scraped and absorbed into future training datasets. This threatens the integrity of the benchmark, as models may achieve high scores by memorizing answers rather than demonstrating genuine reasoning. Preventing this form of benchmark decay remains a significant, field-wide challenge.

\section{\iabs{}}
\subsection{Examples}
We show examples of \iab{} in Table~\ref{tab:imo-answer-bench-examples}.

\begin{table*}[tbh!]
\centering
\resizebox{\linewidth}{!}{%
\begin{tabular}{p{1cm}p{1.6cm}p{7cm}p{9cm}}
\toprule
\textbf{Subj.} & \textbf{Source} & \textbf{Original} & \textbf{Robustified} \\
\midrule
A & Austria MO 2017 & \textit{Determine the maximum $M$ of $x+y+z$ where
$x,y$ and $z$ are positive real numbers with}
\[
16xyz=(x+y)^{2}(x+z)^{2}.
\] & \textit{Let $a, b, c$ be lengths of the sides of some triangle of positive area, satisfying}
\[
    a^2b^2 = 2(a + b - c)(b + c - a)(c + a - b).
\]
\textit{Find the maximum value for $a + b + c$.} \\
\midrule
C & USA TST 2005 & Let $n$ be an integer greater than $1$. For a positive integer $m$, let $S_{m}= \{ 1,2,\ldots, mn\}$. Suppose that there exists a $2n$-element set $T$ such that
(a) each element of $T$ is an $m$-element subset of $S_{m}$;
(b) each pair of elements of $T$ shares at most one common element;
and
(c) each element of $S_{m}$ is contained in exactly two elements of $T$.
Determine the maximum possible value of $m$ in terms of $n$. & For a positive integer $m$, let $S_{m}= \{ 1,2,\ldots, 25m\}$. Suppose that there exists a $\underbrace{\text{$50$-element}}_{\text{Modify numerical value}}$ set $T$ such that:
\begin{enumerate}
    \item Each element of $T$ is an $m$-element subset of $S_{m}$; 
    \item Each pair of elements of $T$ shares at most one common element;
    \item Each element of $S_{m}$ is contained in exactly two elements of $T$.
\end{enumerate}
Let $P$ be a set of $50$ random integers.
Suppose we define a function $\underbrace{\text{$f(x)=x^2+2x+1$}}_{\text{Add distractors}}$.
Determine the maximum possible value of $m$. \\ 
\midrule
G & USA TST 2024 & Let $ABC$ be a triangle with incenter $I$. Let segment $AI$ intersect the incircle of triangle $ABC$ at point $D$. Suppose that line $BD$ is perpendicular to line $AC$. Let $P$ be a point such that $\angle BPA = \angle PAI = 90^\circ$. Point $Q$ lies on segment $BD$ such that the circumcircle of triangle $ABQ$ is tangent to line $BI$. Point $X$ lies on line $PQ$ such that $\angle IAX = \angle XAC$. Prove that $\angle AXP = 45^\circ$.
& Let $ XYZ $ be a triangle with incenter $ J $. Let segment $ XJ $ meets the incircle of triangle $ XYZ $ at point $ K $. Suppose that the angle created by line $ YK $ and line $ XZ $ is $90^\circ$. Let $ R $ be a point such that $ \angle YRX = \angle RXJ = 90^\circ $. Point $ S $ lies on segment $ YK $ such that the circumcircle of triangle $ XYS $ is tangent to line $ YJ $. Point $ T $ lies on line $ RS $ such that $ \angle JXT = \angle TXZ $. Let $\gamma$ be the value of $\angle XTR$ in terms of degree, $\underbrace{\text{compute} \ \frac{\gamma}{3}}_{\text{compute instead prove}}$.
\\
\bottomrule
N & Czech-Slovak Math Olympiad 2017  &Let $k\neq0$ be an integer and suppose that there the number of ordered
pairs $(x,y)$ of integers satisfying
\[
k=\frac{x^{2}-xy+2y^{2}}{x+y}
\]
is odd. Find all possible values of $k$. & Find all even integers $d$ such that the number of ordered integer pairs $(x, y)$ satisfying
\[
    \underbrace{(x + 2y - d)^2 = xy}_{\text{substitute $x \leftarrow x + y$, $y \leftarrow k - y$, $d \leftarrow 2k$}}
\]
is even.
\\
\bottomrule

\end{tabular}%
}
\caption{Examples in the \iab{}, per category ({\bf A}lgebra, {\bf C}ombinatorics, {\bf G}eometry, {\bf N}umber Theory).}
\label{tab:imo-answer-bench-examples}
\end{table*}

\subsection{Subject Distribution and Robustification Examples of \iabs{}}

At the IMO, the problems are typically classified into four main categories: Algebra, Combinatorics, Geometry and Number Theory. Therefore, we also structure our \iab{} in accordance to these four categories as well, where each category has exactly $100$ problems.

\textbf{Algebra} is one of the core competencies for Math Olympiad students and appears at all levels of competitions. Distinct from previous benchmarks \cite{2103.03874}, \ib{} puts more emphasis on Math Olympiad topics, including inequalities, polynomials (including polynomial equations and factorization), functional equations, sequence problems and advanced topics such as Algebraic Number Theory.

\textbf{Combinatorics} problems, despite requiring seemingly basic insights, are notoriously challenging. Successfully solving them serves as a strong indicator of a model’s reasoning capabilities.  The combinatorics set of this benchmark contains problems covering Graph Theory, Enumerative Combinatorics (combinatorial counting problems), Extremal Combinatorics, Existence Combinatorics (problems asking the existence of certain combinatorial objects), Additive Combinatorics, Set Combinatorics, Tiling, Combinatorial Geometry, Operations (problems involving operations, often requiring finding invariant or monovariant properties), and Game Theory. 

\textbf{Geometry} problems at the IMO are well-known for their visual elegance. While there are several existing geometry benchmarks \cite{2103.03874}, they do not cover Math Olympiad level problems.  To address this discrepancy, \ib{} contains geometry problems with short answers spanning subcategories such as angle and sidelength computation, locus problems, and proof-based geometry problems, as well as unconventional categories such as 3D geometry and combinatorial geometry. Additionally, we would like to note that most Math Olympiad level geometry problems are proof-based, and so designing a Math Olympiad level short-answer benchmark for geometry is highly non-trivial.

\textbf{Number Theory} problems typically consist of problems involving objects and properties derived from integers and arithmetic functions, spanning various topics such as Diophantine equations, divisibility problems, polynomials, sequence problems, functional equation problems on the set of integer, existence problems, problems involving arithmetic functions (such as divisor functions, fractional functions), set problems, number theoretic game problems and straategies such as modular analysis, divisor analysis and base representation problems.

These problems serve as a good representation of Math Olympiad problems at various levels and across different national, regional and international contests, as well as the topics covered in these contests. A strong model performance would suggest a high competence level as well as a good knowledge coverage since certain problems can only be solved with a particular problem solving strategy, without which the model would struggle to provide a rigorous with the correct answer.

\subsection{Effects of robustification} \label{appendix:additional_results}
To examine the effect of robustification for \iab{}, we also evaluate on the original, unmodified problems and present the results in Table \ref{tab:imo-answer-bench-original-result}. The models perform significantly better on the original problems, where the gap could be as high as \textbf{$11.2\%$} for \ofourmini{}. This indicates that our robustification effort does create a significant challenge for the models.

\begin{table*}[h!] 
\centering
\label{tab:math-comparison}
\resizebox{\linewidth}{!}{%
\begin{tabular}{l|cccc|c}
\toprule
\textbf{Model} & \textbf{Algebra} & \textbf{Combinatorics} & \textbf{Geometry} & \textbf{Number Theory} & \textbf{Overall accuracy} \\
\midrule
\textbf{\grok{} (Original)} & \bf{78.8\%} & \bf{61.8\%} & 81.4\% & 78.3\% & \textbf{75.0\%} \textcolor{blue}{\scriptsize (+1.9)} \\
\grok{} (Robustified) & 75.5\% & 55.9\% & 80.1\% & 80.9\% & 73.1\% \\
\midrule
\textbf{\gemini{} (Original)} & 77.8\% & 53.0\% & 77.4\% & \bf{78.8\%} & \textbf{71.7}\% \textcolor{blue}{\scriptsize (+3.5)} \\
\gemini{} (Robustified) & 73.4\% & 48.0\% & 74.2\% & 77.1\% & 68.2\% \\
\midrule
\textbf{\ofourmini{} (Original)} & 75.1\% & 52.9\% & \bf{82.5\%} & 75.1\% & \textbf{71.4}\% \textcolor{blue}{\scriptsize (+3.5)} \\
\ofourmini{} (Robustified) & 71.2\% & 46.6\% & 78.4\% & 75.3\% & 67.9\% \\
\midrule
\textbf{\othree{}(Original)} & 67.4\% & 46.8\% & 74.1\% & 67.5\% & \textbf{63.9}\% \textcolor{blue}{\scriptsize  (+2.8)} \\
\othree{} (Robustified) & 62.8\% & 43.0\% & 70.6\% & 68.0\% & 61.1\% \\
\midrule
\textbf{\sonnet{} (Original)} & 28.2\% & 15.5\% & 27.6\% & 27.6\% & \textbf{24.8}\% \textcolor{blue}{\scriptsize (+1.8)} \\
\sonnet{} (Robustified) & 20.6\% & 17.8\% & 26.0\% & 27.6\% & 23.0\% \\
\bottomrule
\end{tabular}%
}
\caption{Comparison between \iab{} results (Robustified) and results for \iab{} before robustification (Original). Results are averaged over 8 samples.}
\label{tab:imo-answer-bench-original-result}
\end{table*}

\subsection{Towards Consistent Problem Statements and Answer Evaluation}
\label{sec:consistent-eval}
Another common issue with language models solving complex Math Olympiad problems is that these models often misinterpret the statement of such problems, or the problem formulation leads the models to produce unintended outputs. Thus, we employ several additional strategies on top of robustification to ensure that the models can interpret the problems properly as follows.
\begin{itemize}
    \item Instead of asking for a series of numbers satisfying certain conditions (which is hard to verify), we instead reformulate the problem so that its answer is a unique number that is the sum or some other non-trivial function of many inputs.
    \item Simplifying the answer as much as possible to avoid confusion.
    \item Being more specific with the problem statement to excuse possible issues with special characters, such as angle degrees in geometry problems.
    \item Avoiding questions with binary answers (yes/no), such as existence questions (which are extremely common in Math Olympiad contests), as they can be guessed without solving the problem or proving the result rigorously. Instead, we will reformulate the problem in such a way that it would produce a non-trivial answer.
\end{itemize}

\subsubsection{Ensuring unique non-trivial answer}
\label{ssec: design-non-trivial-answer}
\paragraph{Example 1} In this example, instead of asking the model to characterize all such numbers $m$, we ask the model to compute a certain expression, which results in $1012$, a value that the model is unlikely to guess by mere chance.

\textit{Original problem}: "For a positive integer $m$, let $a_1, a_2, \ldots, a_{m+1}$ satisfy $3^i < a_i < 3^{i+1}$ for each $i$. Find the maximum and minimum possible values of \begin{align} \sum_{1\leqslant x\leqslant m+1}\prod_{y\neq x}\frac{a_{x}a_{y}-1}{a_{x}-a_{y}}. \end{align}"

\textit{Original answer}: “maximum of $0$ and minimum of $0$ if $m$ is odd, and maximum of $1$ and minimum of $1$ if $m$ is even.“

\textit{Modified problem}: “For a positive integer $m$, let $a_1, a_2, \ldots, a_{m+1}$ satisfy $3^i < a_i < 3^{i+1}$ for each $i$. Let \begin{align} A_m = \sum_{1\leqslant x\leqslant m+1}\prod_{y\neq x}\frac{a_{x}a_{y}-1}{a_{x}-a_{y}}. \end{align} Find $\sum_{i=1}^{2025} A_m^2$”

\textit{Modified answer}: “1012”

\paragraph{Example 2} In this example, instead of asking the model to characterize all solution tuples, which can be hard to evaluate in the natural language form, we ask the models to compute the sum of the elements.

\textit{Original problem}: “Let $a_1, a_2, \ldots, a_{2025}$ be positive integers such that for each positive integer $m$,

$$\left(\left (\sum^{2025}_{j=1} j a^n_j \right)-1\right)^{\frac{1}{n+1}}$$

is an integer. Find all possible $(a_1, a_2, \ldots, a_{2025})$.”

\textit{Original answer}: “$\left(a_{1},\ldots,a_{2025}\right)=(1,k,\ldots,k)$ with $k=2+3+\cdots+2025=2051324$”

\textit{Modified problem}: “Let $a_1, a_2, \ldots, a_{2025}$ be positive integers such that for each positive integer $m$,

$$\left(\left (\sum^{2025}_{j=1} j a^n_j \right)-1\right)^{\frac{1}{n+1}}$$

is an integer. Find all possible values of $a_1+a_2+ \cdots +a_{2025}$.”

\textit{Modified answer}: “4151879777”

\paragraph{Example 3} In this example, instead of asking the model to characterize all such numbers $m$, we ask the models to \emph{count} the number of such numbers in a certain range, which results in $1009$, a value that the model is unlikely to guess by mere chance.

\textit{Original problem}: “Find all positive integers $m \geq 2$ that satisfy the following condition: For any $m$ distinct positive integers $(n_1, \ldots, n_m)$, at least one of the following two conditions holds: $n_1 + \ldots + n_m$ is a multiple of $m$, or there exists a permutation $(k_1, \ldots, k_m)$ such that $k_1 + 2k_2 + \ldots + mk_m$ is a multiple of $m$.”

\textit{Original answer}: “All powers of 2 and all odd numbers”

\textit{Modified problem}: “Find the number of all positive integers $2\leq m \leq 2000$ that satisfy the following condition: For any $m$ distinct positive integers $(n_1, \ldots, n_m)$, at least one of the following two conditions holds: $n_1 + \ldots + n_m$ is a multiple of $m$, or there exists a permutation $(k_1, \ldots, k_m)$ such that $k_1 + 2k_2 + \ldots + mk_m$ is a multiple of $m$.”

\textit{Modified answer}: “1009”

\subsubsection{Answer simplification}
\label{ssec: answer-simplification}
\paragraph{Example} In the example below the original answer mixes notations and adds a potentially confusing quantifier, so we simplify it.

\textit{Original Problem}: “Let $P$ be a function from the set of integers to itself such that for all integers $h, m$, $P^{h^2 + m^2}(h+m-1) = mP(m-1) + hP(h-1) + (h+m-1)$. Find all possible functions $P$.”

\textit{Original answer}: “$P\equiv -1$ or $P(x)=x+1$ for all $x\in\mathbb{Z}$.”

\textit{Modified/simplified answer}: “$P(x)=-1, P(x)=x+1$”

\subsubsection{Handling geometric quantities}
\label{ssec: geometry-example}
\paragraph{Example} For geometry, if the model is asked to find an angle, we make sure to include “in degrees” in the problem statement. This prevents problems mixing radians and degrees and model misinterpretations of special characters marking degrees such as “\^{}o” or “\^{} \{$\backslash$ circ\}”.

\subsubsection{Reformulating questions with binary answers}
\label{ssec: handling-binary-answers}

\paragraph{Example} Below is an interesting example where the problem is very difficult but the answer is a binary yes/no, which can be guessed without solving the problem. Therefore, instead of asking the model to show existence, we ask the model to find the smallest positive integer to satisfy such a property, which retains the difficulty level while preventing the model from guessing the answer.

\textit{Original problem}: “Is there a positive integer $n$ such that $\frac{(a+b)(b+c)(c+a)+a+b+c}{abc} = n$ for infinitely many integer triples $(a,b,c)$?”

\textit{Original answer}: “It exists”

\textit{Modified problem}: “Find the smallest positive integer $n$ such that there exists infinitely many triple $(a,b,c)$ of distinct positive integers such that $\frac{(a+b)(b+c)(c+a)+a+b+c}{4abc} = n$.”

\textit{Modified answer}: “3”

\subsection{Query prompt for \ags{}} \label{subsec:answergrader-prompt}
The following prompt was used to query the \ag{} for \iab{}
\begin{quote}
\ttfamily 

\noindent\textbf{\# System Role: Deterministic Mathematical Autograder}

You are a precise, automated grading system. Your sole function is to determine if the final answer provided in the \texttt{Model Solution} is mathematically equivalent to the \texttt{Golden Answer}. You must NOT grade the reasoning or steps, only the final result.

\medskip 
\noindent\textbf{\# 1. Grading Guidelines (Equivalence Rules)}

Equivalence is mandatory for a correct grade. You must rigorously verify if the answers represent the exact same mathematical value or expression, even if the format differs.
\begin{itemize}[label=*, leftmargin=*, nosep]
    \item \textbf{**Algebraic Equivalence:**} e.g., `n(n+1)/2` is equivalent to `n\textasciicircum{}2/2 + n/2`. You must verify the algebra.
    \item \textbf{**Numerical Equivalence:**} e.g., `1/2` is equivalent to `0.5`; `sqrt(2)/2` is equivalent to `1/sqrt(2)`.
    \item \textbf{**Set/List Equivalence:**} Unless specified as an ordered tuple/vector, the order of elements does not matter (e.g., \{1, 2\} is equivalent to \{2, 1\}).
    \item \textbf{**Partial Credit:**} No partial credit is allowed. If the answer is incomplete or partially incorrect, it is incorrect.
    \item \textbf{**No Answers:**} If no clear, unambiguous final answer can be extracted, the solution must be graded as incorrect.
\end{itemize}

\medskip
\noindent\textbf{\# 3. Output Protocol (Strict Compliance Required)}

You must execute the task using a two-part structure. Failure to follow this structure will result in task failure.

\medskip
\noindent\textbf{**Part 1: Analysis (Chain-of-Thought)**} \\
You MUST perform your analysis within \textless{}thinking\textgreater{}\textless{}/thinking\textgreater{} tags. Make your thinking concise. This section details your reasoning process and must follow these steps sequentially:
\begin{enumerate}[leftmargin=*, nosep]
    \item \textbf{**Golden Answer:**} State the Golden Answer.
    \item \textbf{**Extracted Model Answer:**} State the extracted answer based on the Extraction Protocol. If none found, state "No clear final answer found."
    \item \textbf{**Equivalence Analysis:**} Compare the two answers using the Grading Guidelines. Detail the steps taken to verify mathematical equivalence (e.g., simplification, algebraic manipulation). You must actively try to prove they are the same before concluding they are different.
    \item \textbf{**Conclusion:**} State the final determination ("Correct" or "Incorrect").
\end{enumerate}

\medskip
\noindent\textbf{**Part 2: Final Grade**} \\
Immediately following the closing \textless{}/thinking\textgreater{} tag, output \textbf{**ONLY**} the final grade.
\begin{itemize}[label=*, leftmargin=*, nosep]
    \item If Correct: \textbackslash{}boxed\{Correct\}
    \item If Incorrect: \textbackslash{}boxed\{Incorrect\}
\end{itemize}

\medskip
\noindent\textbf{**CRITICAL CONSTRAINT: Do not add any text, explanations, or formatting outside the \textless{}thinking\textgreater{} tags or the final \textbackslash{}boxed\{\} output.**}

\medskip
\hrule
\medskip
\noindent\textbf{Output exmaple:}

\textless{}thinking\textgreater{}
\begin{enumerate}[label=\arabic*., leftmargin=*, topsep=2pt, itemsep=2pt]
    \item \textbf{**Golden Answer:**} $(-\infty, -4) \cup (-4, \infty)$
    \item \textbf{**Extracted Model Answer:**} $\emptyset$ (the empty set)
    \item \textbf{**Equivalence Analysis:**}
    \begin{quote}
        The Golden Answer is a non-empty set of real numbers.
        The Model Answer is the empty set.
        These two sets are not equivalent. The empty set contains no elements, while the Golden Answer contains an infinite number of elements.
    \end{quote}
    \item \textbf{**Conclusion:**} Incorrect
\end{enumerate}
\textless{}/thinking\textgreater{} \\
\textbackslash{}boxed\{Incorrect\}

\medskip
\hrule
\medskip

\noindent\textbf{\# 4. Input Data} \\
Here is the problem, model solution, and golden answer to grade:

\noindent Problem: \verb|{Problem_Statement}| \\
Model Solution: \verb|{Model_Solution}| \\
Golden Answer: \verb|{Golden_Answer}|

\end{quote}

\section{\ipbs{}}
\subsection{Examples}
We show robustified examples of \ipb{} in Table~\ref{tab:imo2024-robustified-example}.

\begin{table*}[tbh!]
\centering
\resizebox{\linewidth}{!}{%
\begin{tabular}{p{0.15\linewidth} p{0.425\linewidth} p{0.425\linewidth}}
\toprule
\textbf{Source} & \textbf{Original} & \textbf{Robustified} \\
\midrule
IMO '24 P1 & 
Determine all real numbers $\alpha$ such that, for every positive integer $n,$ the integer
$$\lfloor\alpha\rfloor +\lfloor 2\alpha\rfloor +\cdots +\lfloor n\alpha\rfloor$$is a multiple of $n.$ (Note that $\lfloor z\rfloor$ denotes the greatest integer less than or equal to $z.$ For example, $\lfloor -\pi\rfloor =-4$ and $\lfloor 2\rfloor= \lfloor 2.9\rfloor =2.$)
& 
For a real number $r$, let $A(r)$ denote the fractional part of $2r$ in its decimal representation. For a real number $r$ and a positive integer $n$, define $B(n,r)$ as
 $$
 B(n,r)=\sum_{k=1}^n A(kr).
 $$
 Find all positive real numbers $r$ such that $n(n+1)r - B(n,r)$ is a multiple of $n$ for all positive integers $n$.
\\
\midrule
IMO '24 P2 & 
Determine all pairs $(a, b)$ of positive integers for which there exist positive integers $g$ and $N$ such that
\[ \gcd(a^n + b, b^n + a) = g \]
holds for all integers $n \geq N$. (Note that $\gcd(x, y)$ denotes the greatest common divisor of integers $x$ and $y$.)
& 
For given positive integers $x$ and $y$, we define a sequence $(a_n)_{n \ge 1}$ where $a_n$ is equal to
 \[
 \gcd \left( x^n +y , \, (y-x)\left(\sum_{i=0}^{n-1} y^i x^{n-i-1} - 1\right) \right)
 \]
 for all $n\in \mathbb{N}$. Find all pairs $(x,y)$ of positive integers such that the limit of the sequence $(a_n)$ exists. \\
\midrule
IMO '24 P5 & 
Turbo the snail plays a game on a board with $2024$ rows and $2023$ columns. There are hidden monsters in $2022$ of the cells. Initially, Turbo does not know where any of the monsters are, but he knows that there is exactly one monster in each row except the first row and the last row, and that each column contains at most one monster.

Turbo makes a series of attempts to go from the first row to the last row. On each attempt, he chooses to start on any cell in the first row, then repeatedly moves to an adjacent cell sharing a common side. (He is allowed to return to a previously visited cell.) If he reaches a cell with a monster, his attempt ends and he is transported back to the first row to start a new attempt. The monsters do not move, and Turbo remembers whether or not each cell he has visited contains a monster. If he reaches any cell in the last row, his attempt ends and the game is over.

Determine the minimum value of $n$ for which Turbo has a strategy that guarantees reaching the last row on the $n$-th attempt or earlier, regardless of the locations of the monsters.
& 
On a table of size $3002\times3001$, a stone is placed on the leftmost
 cell of the first row. James and Peter play a game on this table.
 Peter selects $3000$ cells, under the rule that he must choose one
 from each row except the first and last rows (i.e., the $1$st and
 $3002$th row), and there must be at most one selected cell in each column.
 James knows this rule too, but he doesn't know which cells Peter selected. The goal of James is to move the stone to the last row,
 avoiding the cells selected by Peter. The stone can only move to adjacent
 cells on the table. If the stone enters a cell selected by Peter,
 James receives a penalty of 1 point, and the stone returns to its
 initial position (i.e., the leftmost cell). Find the smallest positive
 integer $n$ such that there exists a method for James to achieve
 his goal before receiving a penalty of $n$ points.
\\
\bottomrule
\end{tabular}%
}
\caption{Examples of robustified problems, based on the IMO 2024 competition, for \ipb{}.}
\label{tab:imo2024-robustified-example}
\end{table*}

\subsection{Proof Evaluation Guidelines for \ipbs{}}
\label{sec:proof_evaluation}

In a proof-based problem, the desired conclusion usually is either already given ("Prove that ...") or easy to guess ("Determine with proof whether ..."). Evaluating a solution consists of verifying that each logical step leading to the conclusion is valid. However, grading informal\footnote{i.e. written in natural language, as opposed to a formal language such as LEAN.} proofs contains inherently subjective elements, such as deciding whether a particular claim is justified in sufficient detail. Thus, unlike for short answers, which are either correct or incorrect, it is more appropriate to evaluate proofs on a higher resolution scale, where subjective elements matter less. Additionally, a solution may make partial progress by proving some but not all of the steps of a full solution. It is important to capture this during evaluation.

Traditionally, proof-based Math Olympiad competitions, such as the IMO, score solutions on a 7-point scale. For each problem, a grading rubric outlines how many points are to be awarded for certain partial results. The great majority of solutions receive a polarizing score: either 5-7 points for being essentially correct, or 0-2 points if the problem remains unsolved, generally dictated by specific criteria in the rubric. Although problems often admit multiple solutions, it is rare for a solution to be so novel that it falls completely outside of the rubric (which usually covers the 1-2 most common solution approaches). Thus, despite some elements of subjectivity as mentioned above, scores are typically quite consistent across graders. For further insight into how Math Olympiad grading works, refer to \citet{chen:2023:guidance}.

\subsection{Details of \linyang{}}
\label{appendix:linyang}
We use the exact agentic framework proposed in \citep{huang2025gemini25procapable}, which has been open sourced at \url{https://github.com/lyang36/IMO25} and also contains exact hyperparameters in its binary flags. We used the same thinking budget (32K tokens) per model call as mentioned in the paper.

Given an initial solution, a single pipeline consists of repeated iterations (at most 30) of ``self-verification'' and ``bug-fixing'' on it. Specifically, if the current solution passes self-verification a fixed number (5) of times, then the solution is returned, but if at any time self-verification does not pass, then the model is instructed to observe any mistakes (``bugs'') and fix them, and restart the self-verification process. 

Note that if a consecutive number (10) of verifications fail, then the pipeline exits without a solution. This entire pipeline will be run in parallel multiple times (100) as well, until there is at least one solution returned from any run. Theoretically the model could fail to find any solution after all parallel runs, which occurred for two \ipb{} (Advanced) problems.

\subsection{Common Model Mistakes}
\label{sec:appendix}
Here we list some common mistakes the models were making according to the graders.
\begin{table*}[tbh!]
\centering
\resizebox{\linewidth}{!}{%
\begin{tabular}{p{8cm}|p{12cm}}
\toprule
\textbf{Problem} & \textbf{Model Solution} \\
\midrule
\textbf{PB-Basic-004}: We want to find all strictly increasing functions $\mathbb{R} \rightarrow \mathbb{R}$ such that:
\begin{enumerate}
    \item $g$ is surjective.
    \item $g(g(x)) = g(x) + 20x, \, \forall x \in \mathbb{R}$.
\end{enumerate}

&

First, let us restate the problem clearly:

We want to find all strictly increasing functions $g: \mathbb{R} \rightarrow \mathbb{R}$ such that:  
1) $g$ is surjective,  
2) $g(g(x)) = g(x) + 20x$, for all $x \in \mathbb{R}$.

1. Guess and check for linearity

A common approach for such functional equations is to hypothesize that g might be linear (i.e., of the form $g(x) = ax + b$). Let us try this:

Assume $g(x) = a x + b$, where $a, b \in \mathbb{R}$.

Compute $g(g(x))$:

$g(g(x)) = g(a x + b) = a(a x + b) + b = a^2 x + a b + b.$

According to the functional equation, we need:
$g(g(x)) = g(x) + 20x$.

Hence,  
$a^2 x + a b + b = (a x + b) + 20x = a x + 20x + b$.

Matching coefficients:

• Coefficient of x:  $a^2 = a + 20$.  
• Constant term:      $a b + b = b$.

From the constant term $a b + b = b$, we can factor out $b$:
$b(a + 1) = b  \Rightarrow$  either $b = 0$ or $a + 1 = 1$.

If $a + 1 = 1$, then $a = 0$, but that would force $g(x) = b$, a constant function, which cannot be strictly increasing nor surjective over all $\mathbb{R}$. Hence, that is not a valid solution.

Therefore, we must have b = 0. Now the equation $a^2 = a + 20$ remains. Rewrite it as
$a^2 - a - 20 = 0$.

Factor or use the quadratic formula:
$(a - 5)(a + 4) = 0$.

This gives $a = 5$ or $a = -4$.

2. Strictly increasing and surjective requirement

Since g must be strictly increasing:

• If a = -4, then g would be strictly decreasing, which is not allowed.  
• Hence a must be positive, namely a = 5.

So the only candidate under our linear assumption is
$g(x) = 5x$.

Check surjectivity: The map $x \rightarrow 5x$ is surjective, since for any $y \in \mathbb{R}$, we can choose $x = y/5$ to get $5x = y$. It is also strictly increasing because $5 > 0$.

3. Verification in the original equation

Finally, verify $g(x) = 5x$ satisfies $g(g(x)) = g(x) + 20x$:

• $g(g(x)) = g(5x) = 5(5x) = 25x$.  
• $g(x) + 20x = 5x + 20x = 25x$.

These match, so $g(x) = 5x$ indeed satisfies both the functional equation and the given conditions.

Conclusion

Therefore, the only strictly increasing surjective function $g: \mathbb{R} \rightarrow \mathbb{R}$ satisfying $g(g(x)) = g(x) + 20x$ is  
$g(x) = 5x$.

\\
\midrule


\textbf{PB-Basic-005}: Let $P$ be a polynomial with real coefficients whose leading coefficient is $1$. Suppose that for all nonzero real numbers $x$, we have $P(x) + P(1/x) = \frac{P(x + 1/x) + P(x - 1/x)}{2}$. Determine all possibilities for $P$.

&

We consider polynomials of degree $2$ and $4$.
For degree $2$, we found $P(x) = x^2$ is a solution.
For degree $4$, we found $P(x) = x^4 + bx^2 + cx + 6$. We found $c = 0$. So $P(x) = x^4 + bx^2 + 6$ is a solution for any real number $b$.
We verified that $P(x) = x^2$ and $P(x) = x^4 + bx^2 + 6$ satisfy the given condition.
We showed that polynomials of odd degree are not solutions, and polynomials of degree $0$ and $1$ are not solutions.
We need to show that there are no other solutions.
Assume degree of $P$ is $n$. If $n$ is odd, we showed no solution. So $n$ must be even. Let $n = 2k$.
We have shown that for $n=2$, $P(x) = x^2$ and for $n=4$, $P(x) = x^4 + bx^2 + 6$.

Final Answer: The final answer is $P(x) = x^2 \text{ or } P(x) = x^4 + bx^2 + 6 \text{ for some real number } b$.

\\
\midrule

\end{tabular}%
}
\caption{Examples of common mistakes that the models made for \ipb{}.}
\label{tab:imo-incorrect-answers-examples}
\end{table*}

\subsubsection{Polynomial Assumption}



\begin{problemexample}
\textbf{PB-Basic-004}: We want to find all strictly increasing functions $\mathbb{R} \rightarrow \mathbb{R}$ such that:
\begin{enumerate}
    \item $g$ is surjective.
    \item $g(g(x)) = g(x) + 20x, \, \forall x \in \mathbb{R}$.
\end{enumerate}
\end{problemexample}


A common mistake that language models typically make while solving problems is assuming strong assumptions on the problem without a legitimate or substantial justification for the assumptions. An instance of this behavior is the functional equation problem PB-Basic-004 as shown in Table \ref{tab:imo-incorrect-answers-examples}. The proof of o3 on this problem begins by considering the linear cases. 

"We wish to find all strictly increasing and surjective functions $g : \mathbb{R} \to \mathbb{R}$ satisfying

$g(g(x)) = g(x) + 20x$ for all $x \in \mathbb{R}$.

A natural first step is to check if a linear function works. Suppose

$g(x) = ax + b$

...."

After figuring out $a=5$ and $b=0$, the model claims that it found the unique solution, even though the only cases it checked were when $g$ is linear.

"....

Thus, the unique solution is $g(x) = 5x$."

While the final answer is indeed correct, which a typical short answer benchmark would consider correct, the proof is not rigorous and would get little to no points in a proof-based competition such as the IMO.


\subsubsection{Final Answer Guessing}



\begin{problemexample}
\textbf{PB-Basic-005}: Let $P$ be a polynomial with real coefficients whose leading coefficient is $1$. Suppose that for all nonzero real numbers $x$, we have $P(x) + P(1/x) = \frac{P(x + 1/x) + P(x - 1/x)}{2}$. Determine all possibilities for $P$.
\end{problemexample}


In addition, there are the examples where models try to guess the  final answer by inspecting the cases when the variables are small. They do not try to actually prove why the guessed answer is correct. In the example problem PB-Basic-005, the model does case work with degree $n = 2$ and degree $n = 4$ and guesses the answer is $P(x) = x^2$ and $P(x) = x^4 + ax^2 + b$ without showing these are correct answers (in fact, the correct answer should have been $P(x) = a(x^4+6) + bx^2$) nor that these are all the answers. That being said, the models often can get a lot of correct answers by simply guessing rather than carrying out elaborate derivations to arrive at the correct answer. For more information, we refer the readers to the full example in Table \ref{tab:imo-incorrect-answers-examples}.

\subsubsection{Commonly Missed Easy Problems}

Among many problems that models were not able to solve, we present here the following two pre-IMO difficulty problems from ProofBench-basic.

\begin{problemexample}
    \textbf{PB-Basic-008, (Modified) All-Russia MO 2002}: Let $a,b,c$ be positive reals such that $a+b+c = 1$, prove that $\sqrt{a}+\sqrt{b}+\sqrt{c} \geq 3\sqrt{3}(ab+bc+ca)$.
\end{problemexample}

This problem is a standard symmetric homogeneous inequality in three variables, whose equality condition is $a=b=c$. This is one of the easiest type of inequalities one could encounter in a high school level math competition. However, not a single model we tested got even a partial score on this one.

\begin{problemexample}
    \textbf{PB-Basic-016, (Modified) USAMO 1994 Problem 2}: 101 stones are arranged in a circle, numbered 1 through 101 in order (so that stone 1 is next to stone 101). Each stone is painted either red, white, or blue. Initially, stone 101 is the only blue stone. Among the remaining stones, even-numbered stones are red, and odd-numbered stones are white. We perform a sequence of modifications to the coloring, where in each step, we may choose a stone and repaint it a different color, as long as we ensure that no two adjacent stones are ever the same color. Prove that it is not possible to eventually reach a state where again stone 101 is the only blue stone, but among the remaining stones, all even-numbered stones are white and all odd-numbered stones are red.
\end{problemexample}

This is a standard, easy combinatorics problem that uses the invariance principle. One is required to test various quantities related to the problem, and find an invariant quantity to solve the problem. In this case, the difference between the number of blue followed by white and the number of white followed by blue is invariant. However, no model found any invariant that is related to the problem.

\subsection{Query prompt for \pgs{}} \label{subsec:autograder-prompt}
The following prompt was used to query \pg{} for \ipb{}
\begin{quote}
\ttfamily 

You are an expert grader for the International Mathematics Olympiad (IMO). Your task is to evaluate a proposed solution strictly and rigorously. Keep in mind the standards at the IMO are extremely high: only arguments that are logically sound, complete, and precise should be rewarded.

\medskip
\noindent\textbf{\#\#\# General Scoring Rubric} \\
Scores are assigned on a 0-7 scale. The general guidelines are:
\begin{itemize}[label=*, leftmargin=*, nosep]
    \item \textbf{**7 Points (Correct):**} The solution is complete, correct, and fully rigorous. If the submission contains incorrect attempts or lines of reasoning but ultimately presents a complete and correct solution, it should still be awarded full points; the presence of earlier, discarded work does not detract from the final correct proof.
    \item \textbf{**6 Points (Almost Correct):**} The solution is almost correct with a sound core argument, but contains minor errors in calculation or small gaps in logic. Missing proofs for major components, unjustified claims, or sketchy arguments are \textbf{**not**} eligible for 6 points.
    \item \textbf{**1 Point (Partial Progress):**} The solution demonstrates substantial progress explicitly mentioned in the grading guidelines. Initial observations, reformulating the problem without making substantive headway, or proving partial results not mentioned in the grading guidelines are generally \textbf{**not**} eligible for this score.
    \item \textbf{**0 Points (Incorrect):**} The solution doesn't make substantial progress that is a key step in the full solution or is fundamentally flawed. All partial progress without key results or lacking rigor also fall in this category.
\end{itemize}

\medskip
\noindent\textbf{\#\#\# Input Data and Interpretation} \\
You are provided with the following:
\begin{enumerate}[leftmargin=*, nosep]
    \item \textbf{**Problem Statement:**} The IMO problem.
    \item \textbf{**Ground Truth Solution:**} A reference solution. Assume this solution is correct. It demonstrates one valid approach.
    \item \textbf{**Specific Grading Guidelines:**} Criteria for awarding credit for this specific problem. These guidelines take precedence over the General Scoring Rubric, especially for partial credit.
    \item \textbf{**Proposed Solution:**} The student submission.
\end{enumerate}

\medskip
\noindent\textbf{\#\#\# Evaluation Process} \\
You must follow this structured process:
\begin{enumerate}[leftmargin=*, nosep]
    \item \textbf{**Analyze References:**} Meticulously read and understand the problem and Ground Truth Solution check the Specific Grading Guidelines. Identify the key steps for a complete solution and the criteria for partial credit.
    \item \textbf{**Step-by-Step Verification:**} Verify the logical validity and rigor of every step. Identify all flaws, gaps, assumptions, and errors. \textbf{**Make sure you fully understand every piece of logic behind each step of the proposed solution, you must be careful for solutions that 'pretend' to be correct.**}
    \item \textbf{**Assess Progress:**} Determine the extent of non-trivial progress made.
    \item \textbf{**Score Determination:**} Compare the findings against the Specific Grading Guidelines and the General Rubric to determine the final score.
\end{enumerate}

\medskip
\noindent\textbf{\#\#\# Output Requirements} \\
You must provide your final score in the format \texttt{<points>N out of 7</points>}. Ensure the `\texttt{<points>}` block is used \textbf{**only once**}, as your answer will be parsed based on the first \texttt{<points> </points>} block that appears in your whole response.

\medskip\hrule\medskip

\noindent\textbf{**PROBLEM STATEMENT**} \\
\verb|{problem_statement}|

\medskip
\noindent\textbf{**GROUND-TRUTH SOLUTION**} \\
\verb|{solution}|

\medskip
\noindent\textbf{**SPECIFIC GRADING GUIDELINES**} \\
\verb|{guidelines}|

\medskip
\noindent\textbf{**PROPOSED SOLUTION**} \\
\verb|{student_answer}|

\medskip\hrule\medskip

Present your detailed thought process and formal justification based on the scoring rubric and grading guidelines, and finally present your final score in the format below.

\medskip
\noindent{[Select one of the following options]}

\begin{itemize}[label={}, leftmargin=*, nosep]
    \item \texttt{<points>7 out of 7</points>}
    \item \texttt{<points>6 out of 7</points>}
    \item \texttt{<points>1 out of 7</points>}
    \item \texttt{<points>0 out of 7</points>}
\end{itemize}

\end{quote}

\subsection{Limitations of \pgs{}}
\label{sec:proofgrader-limitations}
\begin{table*}[tbh!]
\centering
\resizebox{\linewidth}{!}{%
\begin{tabular}{p{5cm}|p{15cm}}
\toprule
\textbf{Problem} & \textbf{Model Solution and Grade} \\
\midrule
\textbf{PB-Basic 002}: 

Show that $x^2 + y^2 + z^2 + t^2 \ge xyzt$ for any positive real numbers $x, y, z, t$ that satisfy $2(x + y + z + t) \ge xyzt$.
& 
\textbf{Model Solution}: Of course. This is a great problem that elegantly combines a given condition with a target inequality. The key is to use the given condition to constrain the value of the product $xyzt$.

Here is the step-by-step proof.

**Step 1: Use the given condition and the AM-GM inequality to find a bound for the product $xyzt$.**

We are given two pieces of information for positive real numbers $x, y, z, t$:
1.  The condition: $2(x + y + z + t) \ge xyzt$
2.  The inequality to prove: $x^2 + y^2 + z^2 + t^2 \ge xyzt$

Let's start with the Arithmetic Mean-Geometric Mean (AM-GM) inequality for the four numbers $x, y, z, t$:
$$ \frac{x+y+z+t}{4} \ge \sqrt[4]{xyzt} $$
Multiplying both sides by 4, we get:
$$ x+y+z+t \ge 4\sqrt[4]{xyzt} $$
Now, we can substitute this result into the given condition $2(x + y + z + t) \ge xyzt$:
$$ 2(4\sqrt[4]{xyzt}) \ge xyzt $$
$$ 8\sqrt[4]{xyzt} \ge xyzt $$

$(\ldots)$

\vspace{2mm}

\textbf{Human Grade}:  \textbf{\textcolor{blue}{0 out of 7}}

\vspace{2mm}

\textbf{Autograder Verdict}: \#\#\# Step-by-Step Verification

1.  **Step 1: Bounding the product `xyzt`**
    *   The student starts with the given condition: $2(x + y + z + t) \ge xyzt$.
    *   They correctly apply the AM-GM inequality to $x, y, z, t$: $x+y+z+t \ge 4\sqrt[4]{xyzt}$.
    *   Substituting this into the given condition yields $2(4\sqrt[4]{xyzt}) \ge xyzt$, which simplifies to $8\sqrt[4]{xyzt} \ge xyzt$.

\vspace{2mm}

$(\ldots)$

\vspace{2mm}

The solution is complete, correct, and fully rigorous. It meets all the criteria for a perfect score according to the General Scoring Rubric. It does not contain any errors, minor or otherwise. The method is different from the provided ground truth solution (it's a direct proof rather than a proof by contradiction), but it is equally valid and elegant.

<points>\textbf{\textcolor{red}{7 out of 7}}</points>
\\
\midrule
\textbf{PB-Basic 027}:

Given an acute triangle $ABC$, let $D$ and $E$ be the feet of
 the altitudes from $B$ to $AC$ and $C$ to $AB$, respectively.
 Let $E_{1}$ and $E_{2}$ be the reflections of $E$ with respect
 to $AC$ and $BC$, respectively. If $X$ (not equal to $C$) is an
 intersection point of the circumcircle of $\triangle CE_{1}E_{2}$
 and $AC$, and $O$ is the circumcenter of $\triangle CE_{1}E_{2}$,
 prove that $XO$ is perpendicular to $DE$.
&
\textbf{Model Solution}: 
Let $H$ be the orthocenter of triangle $ABC$. Let $H_a$ and $H_b$ be the reflections of $H$ about $AC$ and $BC$, respectively; then $H_a, H_b$ lie on the circumcircle $(ABC)$. Note that reflecting the line $CH$ about $AC$ gives the line $CH_a$, hence $E_1$ lies on $CH_a$; similarly $E_2$ lies on $CH_b$.

\vspace{2mm}

Key lemma (fixed tangent direction):

Let $l = AC$, $r_1 = CH_a$, $r_2 = CH_b$. For any circle $\omega$ through $C$ intersecting $l, r_1, r_2$ again at $X, P, Q$ respectively, the angle between the tangent to $\omega$ at $X$ and $XC$ is constant (independent of the chosen $P \in r_1$ and $Q \in r_2$). Indeed, in the cyclic quadrilateral $XPCQ$ we have:
\begin{itemize}
    \item $\angle XCP = \angle(l, r_1)$ and $\angle XCQ = \angle(l, r_2)$, both constants;
    \item $\angle XPC = \angle XQC$ (equal angles subtending chord $XC$);
    \item hence $2\angle XPC + \angle XCP + \angle XCQ = 180^\circ$, so $\angle XPC = 90^\circ - (\angle(l, r_1)+\angle(l, r_2))/2$ is constant;
    \item by the tangent--chord theorem, this $\angle XPC$ equals the angle between the tangent at $X$ and $XC$.
\end{itemize}


\vspace{2mm}

($\ldots$)

\vspace{2mm}

\textbf{Human Grade}: \textbf{\textcolor{blue}{6 out of 7}}

\vspace{2mm}

\textbf{Autograder Verdict}:

($\ldots$)

\#\#\# Assessment and Score Determination

The solution's overall structure is elegant, but it hinges entirely on a "Key Lemma" which is false and whose proof is invalid. The argument is fundamentally flawed.

<points>\textbf{\textcolor{red}{0 out of 7}}</points>
\\
\bottomrule
\end{tabular}%
}
\caption{Examples of failure cases of \pg{}.}
\label{tab:proofgrader-bias-examples}
\end{table*}

Despite a high correlation with human grades, \pg{} still has several systematic errors and limitations, including a general tendency to overestimate scores, occasional failure to identify high-level logical errors, and being prone to be overly punitive for minor formatting issues or unconventional yet correct solutions. We demontrate specific examples of these behavior in Table \ref{tab:proofgrader-bias-examples}.  

In PB-Basic 002, the model solution makes a logical error by asserting $ 2(4\sqrt[4]{xyzt}) \ge xyzt $ directly from $ x+y+z+t \ge 4\sqrt[4]{xyzt} $ and $2(x + y + z + t) \ge xyzt $. This comes from an incorrect assumption that if $A \ge B$ and $A \ge C$, then $B \ge C$. Such "specious" errors, while seemingly plausible and easy to overlook without a deep understanding of the problem, are critical and can invalidate an entire solution. \pg{} often fails to identify such deceptive logical inconsistencies.

In PB-Basic 027, the model produces a novel solution entirely different from the established ground truth and grading guidelines. The solution was largely correct, but its 'Key Lemma' omits a critical condition that the segment $PQ$ must have a fixed slope. While the lemma is false as stated, supplying this condition makes its proof an immediate consequence of homothety. Since the rest of the solution is complete, the human grader awarded it 6 out of 7 points. However, because the lemma is technically incorrect, \pg{} marks the entire solution as wrong. This case demonstrates that \pg{} struggles to identify partial progress in solutions not anticipated by the grading guidelines, leading to overly punitive assessments for minor issues.

\section{\igbs{}}
\subsection{Grade distribution for \igbs{}}
\label{sec:grade_distribution}

\begin{figure*}[tbh!]
    \centering
    \resizebox{\linewidth}{!}{%
    \includegraphics{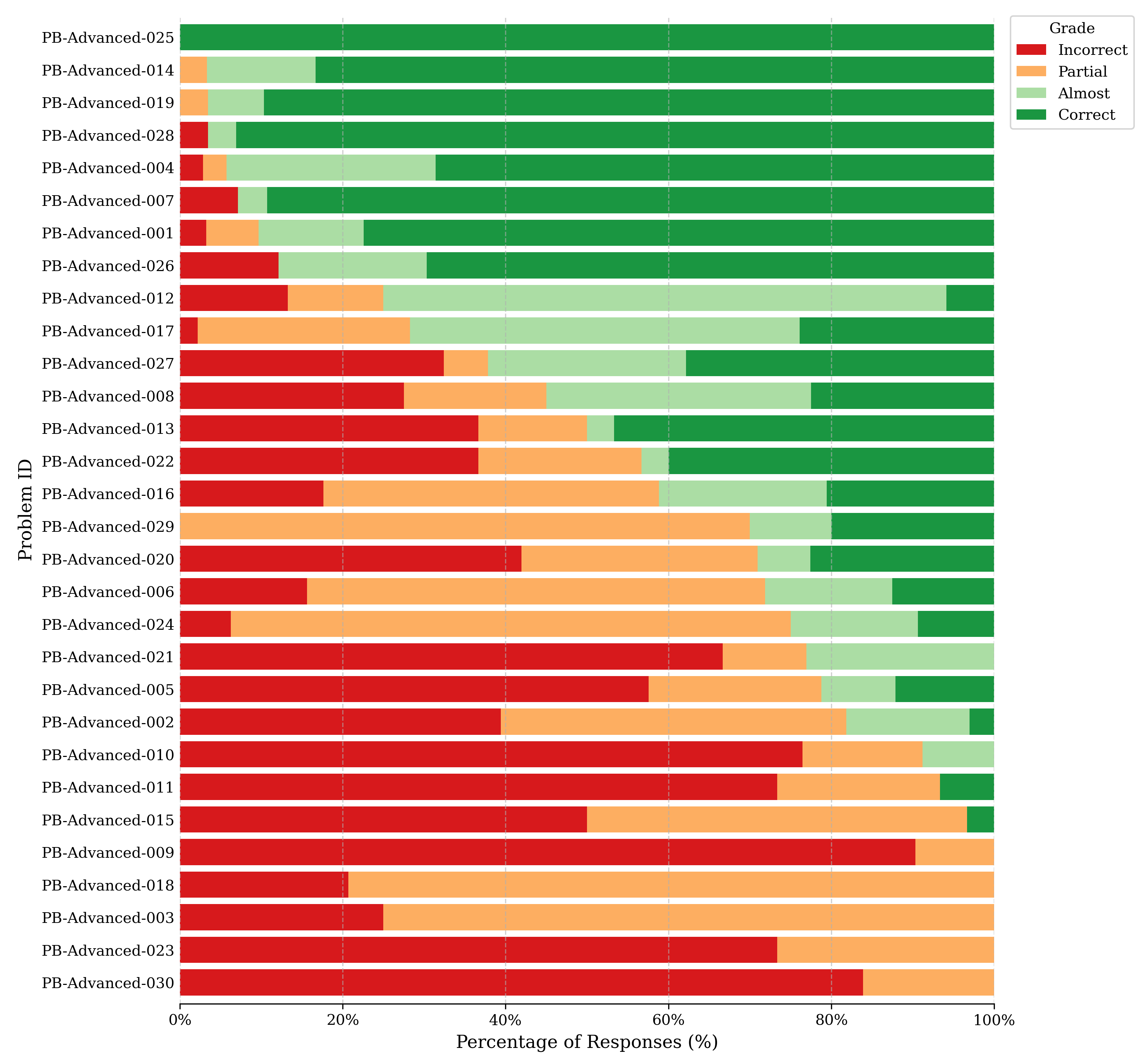}
    }
    \caption{Grade distribution across examples in \igb{}}
    \label{fig:full-difficulty}
\end{figure*}

This section presents the human-assigned grade distribution for the \igb{} benchmark. As shown in Figure~\ref{fig:full-difficulty}, the aggregate count of correct versus incorrect grades across the entire dataset is balanced.

However, the distribution of grades (correct, almost, partial, incorrect) is not uniform on a per-problem basis. This variance is expected as it reflects the natural distribution of scores that proof- evaluation models will encounter in grading solutions, as problems inherently differ in difficulty.

\subsection{Query Prompt} 
\label{app:grading_prompts}

This section details the prompts used for the three evaluation settings in \igb{}. A common definition of the scoring criteria is used across all settings, inserted into the prompts as indicated by \texttt{\{SCORING\_CRITERIA\}}.

\subsection{Grader Prompt}
The following prompt was used for the vanilla setting:
\begin{quote}
Carefully analyze the given problem statement and the proposed solution, and
then write out your analysis regarding the correctness of the proposed solution.

After the analysis, you must provide a score based on the following criteria:

\begin{itemize}
    \item \textbf{incorrect:} The solution is completely incorrect or irrelevant.
    \item \textbf{partial:} The solution is partially correct but has significant errors or omissions.
    \item \textbf{almost:} The solution is almost correct but contains minor errors or inaccuracies.
    \item \textbf{correct:} The solution is fully correct and complete.
\end{itemize}

The very last part of your response must be \textbf{only} one of the following words: incorrect, partial, almost, or correct.

\verb|Problem:{problem}|
\verb|Solution:{solution}|

\end{quote}









\subsection{Label extraction prompt} 
\label{app:grading_extracting_prompt}
The following prompt was used to extract the label from model response for \igb{}. Note that in the majority of cases, the last word of the model (grader) response is one of incorrect, partial, almost, or correct. As a result, we first use python to extract the model grades. We only use prompting to extract the model grades when the last word in the model response is empty or is some different words.

\begin{quote}
\ttfamily 

\noindent\textbf{\#\# Instructions for Extracting Final Scores}

\medskip
\noindent\textbf{**Objective:**} Given an response of an evaluation prompt, extract the final score presented within the response and format it specifically.

\medskip
\noindent\textbf{**Process:**}
\begin{enumerate}[leftmargin=*, nosep]
    \item \textbf{**Analyze the response:**} Scan the response to identify the final score provided by the evaluator.
    \item \textbf{**Extract and format the final answer:**} Present the extracted score on a new line, preceded exactly by "Final answer: ".
\end{enumerate}

\medskip
\noindent\textbf{**Formatting Rules:**}
\begin{itemize}[label=*, leftmargin=*, nosep, itemsep=2pt]
    \item \textbf{**Evaluation Categories:**} The expected output must be one of the following categories: `correct`, `partial`, `almost`, `incorrect`, or `not found`.
    \item \textbf{**Score Identification:**} The extraction is based on identifying the keyword used by the evaluator to summarize their conclusion. The criteria associated with these keywords are:
        \begin{itemize}[label=*, leftmargin=*, nosep, topsep=2pt]
            \item \textbf{**incorrect:**} The evaluator concluded that the solution is completely incorrect or irrelevant.
            \item \textbf{**partial:**} The evaluator concluded that the solution is partially correct but has significant errors or omissions.
            \item \textbf{**almost:**} The evaluator concluded that the solution is almost correct but contains minor errors or inaccuracies.
            \item \textbf{**correct:**} The evaluator concluded that the solution is fully correct and complete.
            \item \textbf{**not\_found:**} The evaluation response does not clearly contain one of the four explicit scores listed above.
        \end{itemize}
    \item \textbf{**Extraction:**} Determine the provided score from the response and extract the category (`correct`, `partial`, `almost`, or `incorrect`). If a score cannot be reliably identified within the text, the output must be `not\_found`.
\end{itemize}

\medskip
\noindent\textbf{**Note:**} No additional markings or explanations are needed beyond "Final answer: " and the extracted answer.

\medskip
Below is the response:

\verb|{Model Response}|
\end{quote}

\end{document}